\documentclass{article}

\usepackage[preprint]{corl_2026} 
\usepackage{graphicx}
\usepackage{hyperref}
\usepackage{booktabs}
\usepackage{subcaption}
\usepackage{multirow}
\usepackage{xcolor}
\usepackage{amssymb}
\usepackage{amsmath}
\definecolor{polina_purple}{rgb}{0.34, 0.16, 0.43}
\definecolor{jeremiah_blue}{rgb}{0.16, 0.34, 0.58}

\usepackage[textsize=tiny]{todonotes}
\usepackage{ifthen}
\newboolean{comments}
\setboolean{comments}{true} 

\ifthenelse{\boolean{comments}}{
  \setlength{\marginparwidth}{2.5cm}
  \newcommand{\zkn}[1]{\todo[color=orange!20, size=\tiny]{Zsolt: #1}}
  \newcommand{\zk}[1]{\textcolor{blue}{Zsolt: #1}}
}{
  \renewcommand{\zk}[1]{}
  \renewcommand{\zkn}[1]{}
}

\title{FlowEval: How Much Adaptation Does an Embodied Foundation Model Need at Deployment?}

\author{
  \mdseries Zhang Xi-Jia\textsuperscript{1}, Roman Yakunin\textsuperscript{1}, Shivang Chopra\textsuperscript{1}, Chengyue Huang\textsuperscript{1}, Che Chen\textsuperscript{2} \\ Siddharth Karamcheti\textsuperscript{1}, Zsolt Kira\textsuperscript{1}\\[5pt]
  Georgia Tech\textsuperscript{1}, University of Michigan\textsuperscript{2}\\
}

\begin{document}
\maketitle

\begin{abstract}
Large-scale pretraining is reducing the amount of adaptation required to deploy embodied foundation models, 
but it is not guaranteed that a pretrained model already provide the level of performance required in every deployment.
Practitioners therefore need to understand how well a pretrained model matches a target deployment and how much adaptation data is left to be collected.
We study whether a small set of successful target-domain demonstrations can reveal whether and how much target-domain data will be required before any adaptation.
We introduce \textsc{FlowEval}, a set of training-free metrics
that use the flow-matching objective to probe an embodied foundation model with a set of demonstrations.
\textsc{FlowEval} quantifies predictive disagreement on successful target demonstrations to assess how well a pretrained model fits a target deployment.
Across four embodied foundation model families: Pi-0.5, MolmoAct2, GR00T, and DreamZero, \textsc{FlowEval} distinguishes seen from unseen deployments across real-world and simulated manipulation.
Further, by finetuning Pi-0.5 and GR00T on 17 RoboCasa365 manipulation tasks, we showed that \textsc{FlowEval} correlates with how success rate improves as additional adaptation data is provided. 
Together, these results suggest that \textsc{FlowEval} can serve as a lightweight, training-free diagnostic of how much model--deployment adaptation may be required.

\end{abstract}

\section{Introduction}

Large-scale pretraining is enabling embodied foundation models to generalize across increasingly broad variation in tasks, environments, and robot embodiments \citep{kim2024openvla, physical_intelligence2025pi05, octo_2023, embodimentcollaboration2025openxembodimentroboticlearning}. Yet broad generalization does not imply that a pretrained model will already provide the level of performance required in every downstream deployment \citep{octo_2023, kim2025finetuningvisionlanguageactionmodelsoptimizing, geminiroboticsteam2025geminirobotics15}. In practice, models may still require target-domain adaptation to acquire novel physical behaviors, transfer to another action space, or improve behaviors that are only partially captured by the pretrained prior. While broad pretraining increasingly reduces the amount of task-specific learning required, adaptation remains an important component of deploying today's embodied foundation models \citep{physical_intelligence2025pi05, fang2026molmoact2, ye2026dreamzero}.

As embodied models become more general-purpose, this changes the deployment problem itself. Rather than asking only how to adapt a pretrained model to a downstream setting, practitioners increasingly need to understand \emph{how much of the required behavior is already provided by pretraining and how much additional adaptation remains}. Importantly, poor zero-shot performance does not reveal whether the desired behavior can be recovered with a small amount of target-domain data or requires substantially more adaptation, and a model with stronger broad capabilities need not require less adaptation for a particular deployment.
This motivates two questions: 
1) whether the target deployment is already seen by the model,
and, ultimately, 2) how much additional data will be required to achieve the desired level of performance?

We study whether the answer to these questions can be probed directly with a small set
of successful target-domain demonstrations. Such trajectories provide examples of the behavior the deployed system is expected to reproduce, and can often be collected with substantially less effort than the full datasets required for finetuning. This suggests a simple hypothesis: downstream adaptation difficulty should depend on how well the pretrained model already fits the intended target behavior. Rather than first finetuning several candidate models and observing which one adapts successfully, we ask whether this model--deployment relationship can be measured \emph{before adaptation}.

A natural quantity to examine is the pretrained model's training loss on successful target-domain trajectories. In language modeling, held-out cross-entropy and perplexity have long been used to characterize how well a model fits a target domain and to select data that better matches that domain \citep{moorelewis2010intelligent,gururangan2020dontstoppretrainingadapt,DoReMi2023}.
Flow matching is particularly well suited to assessing model--data fit because its training objective probes not only the predicted action, but the model's local direction of improvement toward the demonstrated action \citep{lipman2023flowmatchinggenerativemodeling}.
We therefore introduce \textsc{FlowEval}, a set of training-free metrics derived from the flow-matching objective that measure how well a pretrained model fits successful target-domain demonstrations by computing the predictive disagreement. 
\textsc{FlowEval} requires no finetuning and applies whether or not the pretrained model and target deployment share the same action representation.
This is particularly important for deployment, as action representations and deployment interfaces can differ widely across model families and robot embodiments. \citep{embodimentcollaboration2025openxembodimentroboticlearning, gr00tn1_2025}

We evaluate \textsc{FlowEval} on Pi-0.5, MolmoAct2, GR00T, and DreamZero, across real-world and simulated manipulation and multiple robot embodiments. 
Across these settings, \textsc{FlowEval} predicted directional disagreement correlates distinguishes seen deployment distributions from unseen deployment distributions.
More importantly, by finetuning Pi-0.5 and GR00T on 17 RoboCasa365 manipulation tasks, we find that the same signal correlates with how performance scales as additional adaptation data is provided. Together, these results suggest that \textsc{FlowEval} can serve as a training-free diagnostic of how much target-domain adaptation may be required.

\section{Related Work}

\subsection{Adapting embodied foundation models}
\label{sec:related_work_adapting}

\textbf{Downstream finetuning} is the widely adopted approach to adapt a general pretrained policy. State-of-the-art open source embodied foundation models,  including but not limited to DreamZero \citep{ye2026dreamzero}, MolmoAct2 \citep{fang2026molmoact2}, Pi-05 \citep{physical_intelligence2025pi05}, and GR00T \citep{gr00tn1_2025}, provide finetuning pipelines and include finetuning guidelines at release.
Prior to them, OpenVLA \citep{kim2024openvla}, OpenVLA-OFT \citep{kim2025finetuningvisionlanguageactionmodelsoptimizing}, and Octo \cite{octo_2023} expected downstream finetuning even more. On the proprietary end, Generalist Gen1.5 \citep{generalist2026gen15} reported using few gradient step adaptation to better learn novel tasks;  and Gemini Robotics \citep{deepmind2026geminirobotics2,geminiroboticsteam2025geminirobotics15} reported finetuning the model to target embodiments. 
Although Figure AI \citep{figure2026helix02}, Skild \citep{skild2026s1}, and Genesis AI \citep{genesis2026gene265} do not report downstream fine-tuning, notably, their pretraining datasets are directly built for their deployment settings. This is fundamentally different from the situation faced by practitioners, who lack the ability to construct a large-scale pretraining dataset specifically for their deployment.

\textbf{Learning In-Context} is a promising yet constrained approach for a pretrained embodied foundation model to adapt to a new task. 
DreamZero \citep{ye2026dreamzero} and Skild S1 \citep{skild2026s1} provided video as context and shows the model is able to infer tasks from the world to state information in the videos. Generalist Gen1.5 \citep{generalist2026gen15} provided one-shot state to action demonstration as context and shows emerging ability of the model generalizing to novel objects. However, all require the embodiment and action representation to align between the pretrained and deployment settings. \citet{generalist2026gen15} also reported that skills learned in-context are more brittle than finetuned models.

\textbf{Zero-shot} execution is possible in scenarios the embodiment and action representation between pretrained and deployment settings are aligned. 
\citet{fang2026molmoact2,octo_2023,rewardai2026om1} showed zero-shot execution of tasks on supported embodiments. 
Similar to in-context learning, their prerequisites are more restrictive than most practical deployment settings.

\subsection{Training objectives as Evaluations}
\label{sec:related_work_training_objectives}

The optimization objective, such as loss or negative log-likelihood (NLL), is commonly used as an evaluation metric for model performance in machine learning \citep{zhang2017understandingdeeplearningrequires,guo2017calibrationmodernneuralnetworks} and especially on large language models \citep{kaplan2020scalinglawsneurallanguage, hoffmann2022trainingcomputeoptimallargelanguage}. In robotics, validation loss is often used a proxy for performance to select policies \citep{ross2011reductionimitationlearningstructured, zhao2023learningfinegrainedbimanualmanipulation}. As to embodied foundation models including RT-2 \citep{brohan2023rt2}, Octo \citep{octo_2023}, and OpenVLA \citep{kim2024openvla}, however, multiple works \citep{li2024evaluatingrealworldrobotmanipulation,zhou2025autoevalautonomousevaluationgeneralist} showed that validation loss poorly reflects actual robot performance. Interestingly, none of these work \citep{li2024evaluatingrealworldrobotmanipulation,zhou2025autoevalautonomousevaluationgeneralist} tested flow-matching based models.

\textbf{Flow objectives as evaluations}

Likelihood-based flows naturally allows likelihood evaluations. A common use case is to evaluate the model on an entire held-out dataset, using the likelihood to measure how well the model fits unseen data \citep{durkan2019neuralsplineflows,Papamakarios2021normalizing}. In contrast, using the likelihood of a single data point as a measure of whether that point is in- or out-of-distribution is a stronger assumption and is not generally guaranteed \citep{nalisnick2019deepgenerativemodelsknow,han2026conditionalflowmatchingreplace}. 
However, in robotics, where the generated distribution is conditional action likelihoods $p(a\mid s)$, likelihood have been successfully applied as measures of action plausibility or anomaly: \citet{Wenger2025anomalously} use it to detect several types of antagonistic robot behavior, while \citet{NEURIPS2023_zhang_constrainedpolicy,chae2026flowactorcriticofflinereinforcement} effectively applied it inside offline-RL algorithms as an OOD score.

The flow matching loss shares some common properties with likelihood-trained normalizing flows, such as evaluating how well a policy fits a held-out demonstration set. \citet{zha2026laplanguageactionpretrainingenables} evaluates the validation loss of its flow-matching action expert on a held-out split and uses it to compare model scaling. Recent robot policy evaluations also compute the flow matching objective on held-out episodes as an offline action-prediction loss \citep{robocurve2026groot}.
\citet{park2026behavioral} evaluated the relationship between flow-matching loss and success rate, and discovers that lower training loss is associated with higher validation loss and higher performance.

\textsc{FlowEval} differs from prior works in two meaningful ways. (1) We discovered that starting from pure noise provided us with cleaner evaluation signals than noisy actions (Appendix.~\ref{appendix:why-pure-noise})
(2) Instead of using loss to select checkpoints from the same training run, we use pre-adaptation loss to probe how finetuning outcomes vary across target deployments and demonstration budgets.

\section{Methodology}

We begin by clarifying the deployment settings considered in this work. We distinguish the conditions that fall within and outside the scope of our metric evaluation.

\textbf{Demonstration}: We define a demonstration as an observation-state-action trajectory demonstrating successful completion of a task. In this work, we primarily consider demonstrations collected via human teleoperation. Although approaches such as MimicGen \citep{mandlekar2023mimicgendatagenerationscalable}, DemoGen \citep{xue2025demogensyntheticdemonstrationgeneration}, and other forms of augmenting existing demonstrations may be interesting directions to explore, they are beyond the scope of this study (Appendix. ~\ref{appendix:counter-case}).

\textbf{Deployment settings}: We define a deployment setting as one or more manipulation tasks performed under a fixed environment, embodiment, and action representation. We further assume access to a small set of demonstrations for the target task(s). We do not assume that the action representations and dimensions of the deployment setting matches with a model's pretrained distribution, since in actual deployments they often do not.

\textbf{Practical Scope}: Most practitioners do not have access to a foundational model trained specifically for their deployment settings -- only those who pretrain the model themselves do. Instead, they often seek to adapt an existing foundation model to a target deployment without full, or even partial, access to its pretraining data. Their deployment may involve a custom embodiment, camera configuration, task setup, or action representation. Even when these components are otherwise standard, they may still differ from the exact configurations supported by the pretrained model. 
Our definitions therefore describe a deployment scenario that practitioners are most likely to encounter in practice.

\subsection{FlowEval Metrics}
\label{sec:floweval_metrics}

Using flow matching objectives to evaluate how well a trained model fits a given dataset is a promising yet under-explored direction in robotics (Sec. \ref{sec:related_work_training_objectives}). We propose FlowEval metrics based on the intuition that the flow matching objective provides a measurable signal of how well the embodied foundation model fits a deployment dataset, which further affects how much adaptation is needed. 
Since FlowEval relies on properties of flow-based objectives that may not generalize to other training objectives, its metrics are only applicable to embodied foundation models that use flow matching for action prediction.

For vision-language action (VLA) models, the forward pass can be written as
\begin{equation}
    \hat u_A=\Phi_\theta
    \left(A^{\tau_A},h_{vision},h_{language},h_{state},h_{{time}_A}\right),
\end{equation}

A world-action model (WAM) additionally predicts future quantities, which can extend beyond videos. We denote those future quantities by $m$.
\begin{equation}
    \left(\hat u_A,\{\hat u_m\}_m\right)
    =\Psi_\theta\left(
    A^{\tau_A},\{Z_m^{\tau_m}\}_m,
    h_v,h_\ell,h_s,h_{t_A},\{h_{t_m}\}_m
    \right),
\end{equation}

For ground-truth actions $A^*$ and future targets $Z_m^*$, conditional flow matching uses
\begin{equation}
    A^{\tau_A}=(1-\tau_A)A^*+\tau_A\epsilon_A,
    \qquad
    Z_m^{\tau_m}=(1-\tau_m)Z_m^*+\tau_m\epsilon_m,
\end{equation}
with Gaussian noise $\epsilon_A,\epsilon_m\sim\mathcal{N}(0,I)$ and
reference velocities $u_A=\epsilon_A-A^*$ and $u_m=\epsilon_m-Z_m^*$.

FlowEval initializes the action and future-target trajectories from Gaussian noise, setting $A^1=\epsilon_A$ and $Z_m^1=\epsilon_m$. Subsequent states follow the model's inference dynamics. At each inference step $k$, we evaluate $\hat u_{A,k}$ and $\hat u_{m,k}$ against the fixed reference velocities defined above. 
Although validating with held-out trajectories conventionally starts from ground-truth-interpolated noisy actions, we find that starting from pure noise instead yields more stable results, discussed in \ref{appendix:why-pure-noise}.

\paragraph{Velocity error.}
The action and future-prediction velocity errors are
\begin{equation}
    L_{A,k}=\left\|\hat u_{A,k}-u_A\right\|_2^2,
    \qquad
    L_{m,k}=\left\|\hat u_{m,k}-u_m\right\|_2^2,
\end{equation}

These are basically the loss objectives.
We refer to these as velocity error to differentiate from training or validation loss, which conventionally starts from ground-truth-interpolated noisy actions.

\paragraph{Cosine similarity.}
We measure the directional agreements with the reference velocities:

\begin{equation}
\begin{aligned}
    \operatorname{cosine}_{A,k}
    & =
    \frac{
    \left\langle M_A \odot \hat u_{A,k},\; M_A \odot u_A \right\rangle
    }{
    \left\|M_A \odot \hat u_{A,k}\right\|_2
    \left\|M_A \odot u_A\right\|_2
    + \varepsilon
    },\\
    \operatorname{cosine}_{m,k}
    &=\frac{\left\langle\hat u_{m,k},u_m\right\rangle}
    {\left\|\hat u_{m,k}\right\|_2\left\|u_m\right\|_2+\varepsilon}.
\end{aligned}
\end{equation}
where $M_A$ is the action mask, and $\varepsilon>0$ stabilizes the denominator.

\subsection{Design Choices}
\label{sec:design_choices}

\textsc{FlowEval} metrics are defined over the full trajectory of diffusion steps under the model's configuration. However, computing all steps is computationally expensive and may be unnecessary. We conduct an empirical study along two design axes: (i) distinguishability, the extent to which a metric distinguishes different target settings, and (ii) computational efficiency, the cost of computing the metric from target demonstrations. 

We define distinguishability as the ratio of between-task to within-task variance; higher scores indicate that the metric varies more across tasks than across demonstrations within a task. Computational efficiency is naively evaluated by the step index because exact n forward passes are required to compute step n. 
As shown in Table \ref{tab:distinguishability}, \textsc{FlowEval} metrics on Pi-05-Base and DreamZero-DROID achieved distinguishability at step 1 exceeding $90\%$ of the maximum across all steps, suggesting that a single step is empirically sufficient when accounting for computational cost.
We thus only take the first step for future analysis. Full results of distinguishability scores are included in Table \ref{appendix:tab:pi05-full} and \ref{appendix:tab:dreamzero-full}.

\begin{table}[htbp]
\centering
\label{tab:step1_distinguishability}
\begin{tabular}{llrrr}
\toprule
Model & Metric & Step 1 & Best & Relative \\
\midrule
Pi05-Base
    & Cosine & 0.868 & 0.885 & 98.0\% \\
    & Velocity Error     & 1.097 & 1.104 & 99.4\% \\
\midrule
DreamZero-DROID
    & Action Cosine & 0.455 & 0.498 & 91.3\% \\
    & Action Velocity Error     & 0.244 & 0.267 & 91.3\% \\
    & Video Cosine  & 0.799 & 1.170 & 68.3\% \\
    & Video Velocity Error      & 0.735 & 1.141 & 64.4\% \\
\bottomrule
\end{tabular}
\vspace{0.5em}
\caption{Distinguishability score at step 1 compared with the best value observed across all steps. Relative performance is computed as $100 \times$ step 1 / Best. Evaluated on Robocasa365 target set.}
\label{tab:distinguishability}
\end{table}

\vspace{-0.5em}
\section{Experiments}

\subsection{Predicting Unseen Deployments}
\label{sec:ood}

We first test whether \textsc{FlowEval} reflects how well a deployment's required behavior is supported by the pretrained model. Our hypothesis is that successful demonstrations from a seen deployment should induce less predictive disagreement than demonstrations from an unseen deployment. Although scores may be noisy for individual trajectories, we expect this difference to emerge at the distribution level.

We test this hypothesis across three vision-language-action models: Pi-05, GR00T, MolmoAct2, and one world-action model: DreamZero, on deployment settings whose relationship to pretraining is known. 
For each model, we compare demonstrations from a deployment setting represented during pretraining (\emph{seen}) with demonstrations from a deployment setting absent from pretraining (\emph{unseen}).

\begin{table}[b]
\centering
\begin{tabular}{lccccccc}
\hline
In-Pretraining & DROID & Robocasa & YAM & LIBERO & Roboset & OpenArm & Franka \\
\hline
Pi-05          & \textcolor{gray!100}{No} & No & \textcolor{gray!100}{No} & \textcolor{gray!100}{No} & \textcolor{gray!100}{No} & No & No \\
GR00T-N1.6     & Yes & Yes & \textcolor{gray!100}{No} & \textcolor{gray!100}{No} & \textcolor{gray!100}{No} & No & No \\
MolmoAct2      & Yes & No & Yes & Yes & No & No & No \\
DreamZero      & Yes & No & No & No & No & No & No \\
\hline
\end{tabular}
\vspace{0.5em}
\caption{The full set of model–dataset pairs on which FlowEval metrics are evaluated, along with whether each dataset was included in model pretraining. Information about those datasets is based on model releases \citep{physical_intelligence2025pi05, gr00tn1_2025, fang2026molmoact2, ye2026dreamzero} and, where applicable, company blogs \citep{pi05-blog, gr00tn16-blog, dreamzero-blog}. `\textcolor{gray!100}{No}' indicates that we found no evidence that the dataset was included in pretraining, although the authors did not explicitly state that it was excluded. `No' indicates that the authors explicitly stated that the dataset was excluded, or the pretraining dataset is publicly released and does not contain that dataset.}
\end{table}

\textbf{Datasets.} We compute FlowEval on 7 datasets across real-world and simulation, multiple action representation spaces, and three embodiments: Franka, Bimanual OpenArm, and Bimanual YAM. We collected the Franka and OpenArm datasets ourselves, and additionally use 5 internet datasets: DROID \citep{khazatsky2025droid}, Robocasa365 \citep{nasiriany2026robocasa365largescalesimulationframework}, MolmoAct2-Bimanual-YAM \citep{fang2026molmoact2}, LIBERO \citep{liu2023liberobenchmarkingknowledgetransfer}, and 
RoboSet \citep{bharadhwaj2023roboagentgeneralizationefficiencyrobot}. The exact action representation and checkpoint configurations are discussed in Appendix \ref{appendix:sec:ood}.
Note that \textsc{FlowEval} can be computed between different action representations. 
However, doing so confounds unseen deployment with a change in action representation alone and does not test the other factors of interest. We therefore restrict the main analysis to model--deployment pairs with matching action spaces between pretraining and evaluation. We additionally report cross-representation results in Appendix \ref{appendix:sec:ood}. As expected, these pairs exhibit significantly higher velocity error and lower cosine similarity than pairs from seen deployments.

\begin{figure*}[h]
    \centering

    \begin{minipage}{0.49\textwidth}
    \subcaption{Pi-05, Delta Cartesian}
        \centering
        \includegraphics[width=\linewidth]{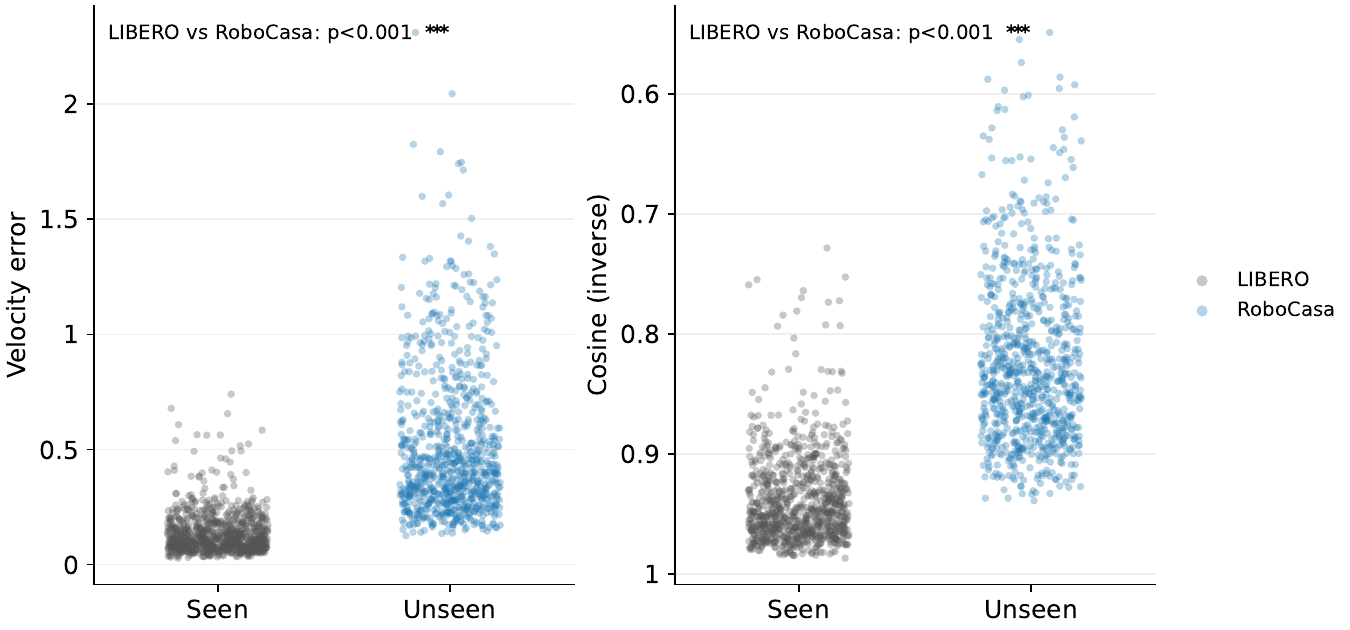}
    \end{minipage}
    \hfill
    \begin{minipage}{0.49\textwidth}
    \subcaption{MolmoAct2, Delta Cartesian}
        \centering
        \includegraphics[width=\linewidth]{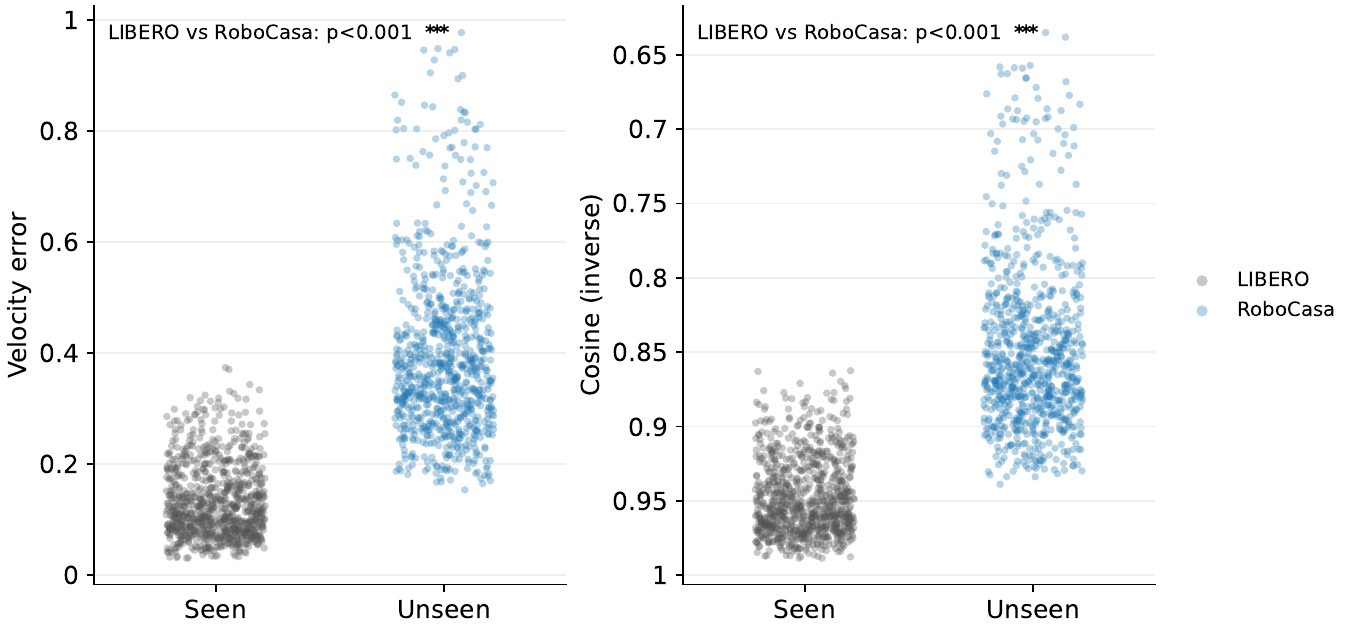}
    \end{minipage}

    \vspace{0.5em}

    \vspace{0.5em}

    \begin{minipage}{0.49\textwidth}
    \subcaption{GR00T-N1.6, Relative Joint-Position}
        \centering
        \includegraphics[width=\linewidth]{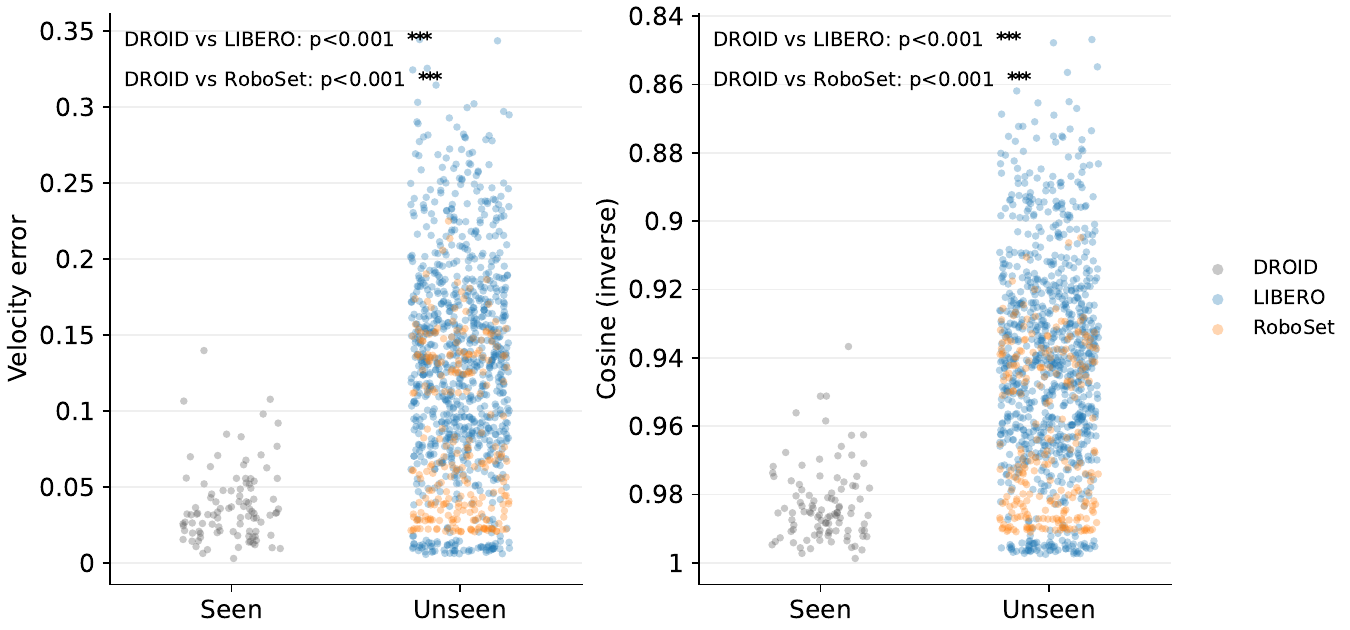}
    \end{minipage}
    \hfill
    \begin{minipage}{0.49\textwidth}
    \subcaption{DreamZero, Relative Joint-Position}
        \centering
        \includegraphics[width=\linewidth]{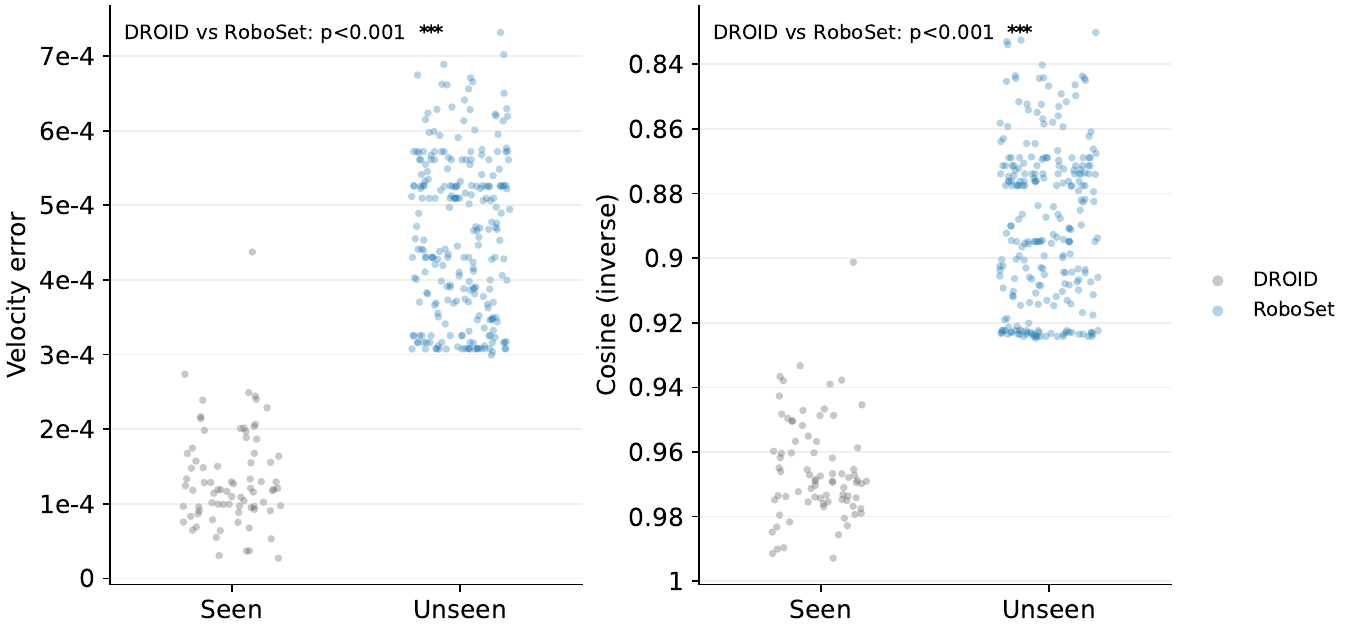}
    \end{minipage}

    \caption{FlowEval results across various models and action representations.}
    \label{fig:ood}
\end{figure*}

We test whether the seen and unseen deployments are separated using a one-sided two-sample Kolmogorov--Smirnov test. For velocity error, the alternative tests whether target distribution lean toward larger errors; for cosine similarity, it tests whether target distribution lean toward smaller similarities. Across Pi-05, GR00T, MolmoAct2, and DreamZero, the corresponding distributions are significantly separated in this direction. The separation is also visible at the distribution level: trajectories from the pretraining setting extend closer to the limiting values of zero velocity error and unit cosine similarity, whereas target trajectories do not reach the same favorable tail.

As a control, we evaluate new demonstrations excluded from pretraining but collected under the same seen settings. \textsc{FlowEval} scores of these demonstrations cluster with the seen deployment and remain distinct from the unseen deployments (Fig. \ref{appendix:fig:part1-additional}).
Because RoboCasa is the only public in-pretraining dataset that provides such held-out trajectories, this control is limited to Robocasa (Appendix \ref{appendix:sec:ood}). Together, these results support FlowEval’s ability to identify unseen deployments and provide preliminary evidence that it recognizes seen deployments from new demonstrations.

\subsection{Predicting downstream adaptation}
\label{sec:regression}

We show that \textsc{FlowEval} metrics are capable of predicting how performance scales with data under deployment shift. To determine whether the scaling behavior simply comes from more data or is specifically conditioned on deployment shift, we experiment with two groups of finetuning where (1) the deployment settings is genuinely absent from the model's pretraining (2) the deployment settings is included in the model's pretraining, but new demonstrations for the settings are available.

We use the Robocasa365 dataset \citep{nasiriany2026robocasa365}, which is built for studying downstream finetuning behavior and thus provide us with cleanly separated pretraining and held-out datasets. Among widely used robot manipulation benchmarks \citep{liu2023libero, mees2022calvinbenchmarklanguageconditionedpolicy, li2024evaluatingrealworldrobotmanipulation, li2024behavior1k, james2019rlbenchrobotlearningbenchmark, zhang2024vlabenchlargescalebenchmarklanguageconditioned}, RoboCasa365 uniquely provides the scale and a pretraining/target split over the same task set that this comparison requires, without confounding deployment shift with differences in semantic-level planning.
We use all shared atomic-seen manipulation tasks between the pretrain set and target set, so that we are able to finetune on held-out demonstrations. For deployment settings absent from pretraining, we finetune Pi-05 base and GR00T-N1.5, neither of which was pretrained on RoboCasa or RoboCasa365. For settings represented in pretraining, we finetune Pi-05-RoboCasa \citep{nasiriany2026robocasa365} and GR00T-N1.6, both of which were pretrained on RoboCasa365.

\begin{figure}[htbp]
    \centering
    \begin{minipage}{0.49\textwidth}
    \subcaption{Pi-05}
        \centering
        \includegraphics[width=\linewidth]{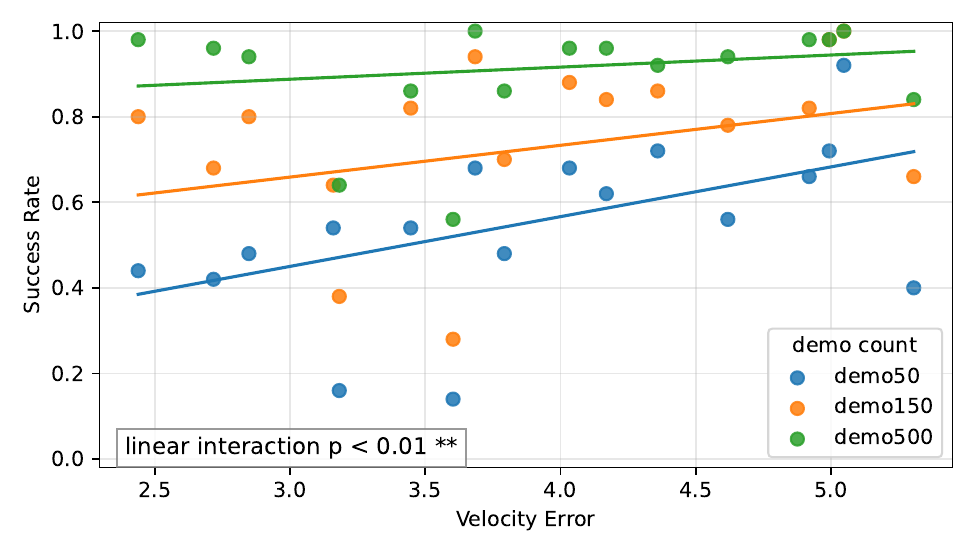}
    \end{minipage}
    \hfill
    \begin{minipage}{0.49\textwidth}
    \subcaption{GR00T-N1.5}
        \centering
        \includegraphics[width=\linewidth]{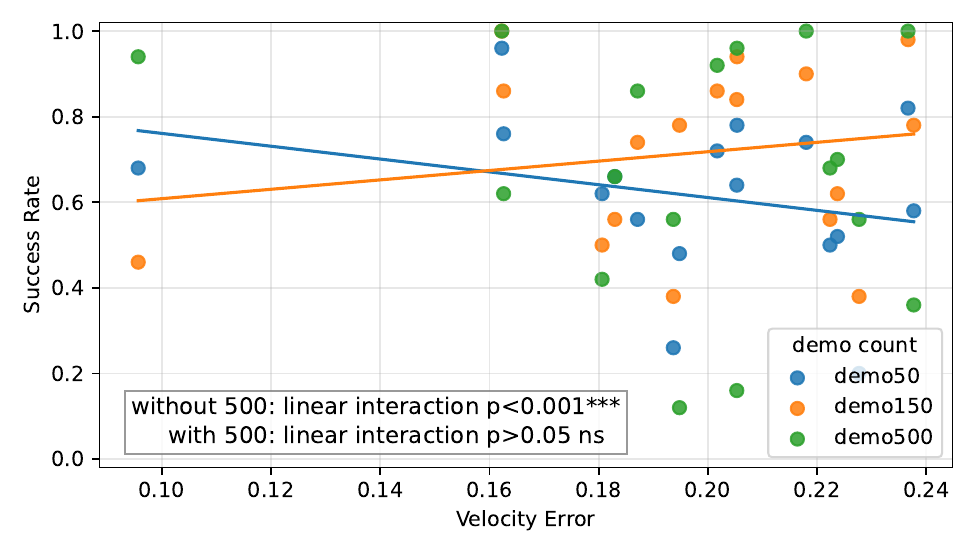}
    \end{minipage}

    \begin{minipage}{0.49\textwidth}
        \centering
        \includegraphics[width=\linewidth]{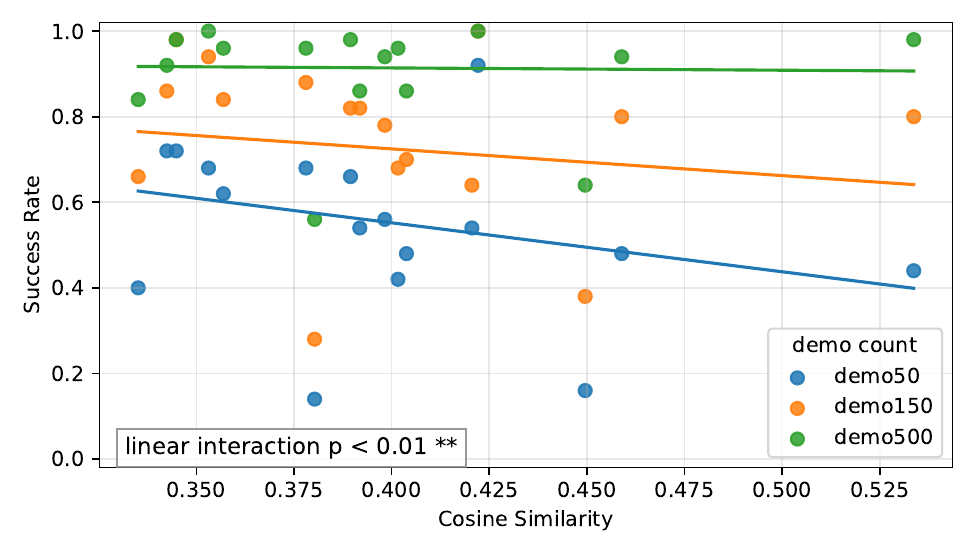}
    \end{minipage}
    \hfill
    \begin{minipage}{0.49\textwidth}
        \centering
        \includegraphics[width=\linewidth]{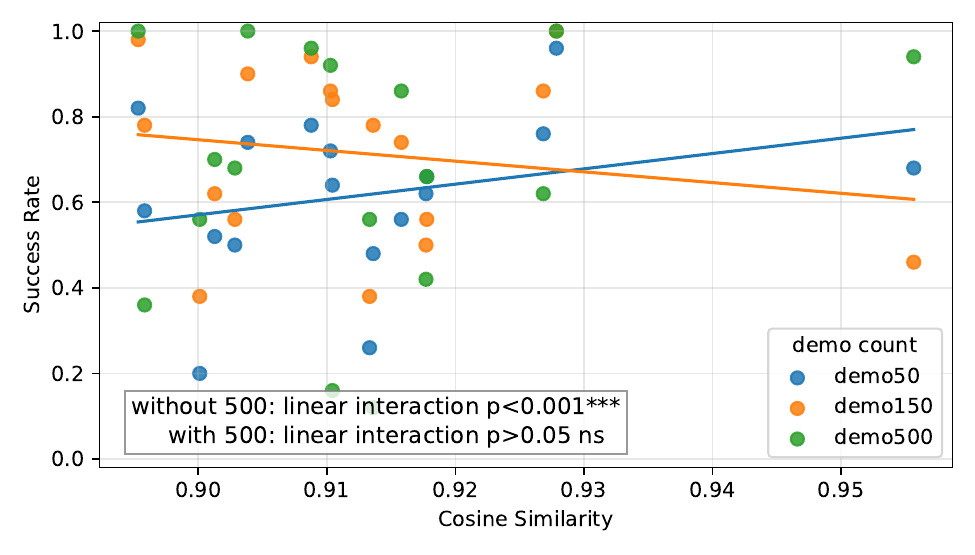}
    \end{minipage}
    \caption{Regression between success rate and metric values, with significances displayed in the plots. All interaction effects are significant. Under the GR00T-N1.5 500-demonstration condition, training instability causes 29.4\% of runs to collapse; these runs are excluded from the analysis (Appendix~\ref{appendix:sec:collapse}). Training collapse is negligible in other conditions.}
    \label{fig:regression}
\end{figure}
\vspace{-0.4em}

For each task, we compute \textsc{FlowEval} on the pretrained model before target-domain adaptation, and then finetune with increasing amounts of target data. Following RoboCasa365 \citep{nasiriany2026robocasa365}, we use 50, 150, and 500 demonstrations per task. Each regression observation corresponds to one task--demonstration-count pair, with success rate given by the highest observed checkpoint performance during that finetuning run and the task-level \textsc{FlowEval} score fixed across data budgets. This design tests whether the initial model--domain relationship measured by \textsc{FlowEval} predicts differences in the response to additional adaptation data. We fit the following model:
\vspace{-0.7em}

$$
R=\beta_0
+\beta_1 L
+\beta_2 m
+\beta_3 Lm
+\epsilon,
$$

where
$
L=\log(N)-\overline{\log(N)}
$
is the centered log number of demonstrations and
$ m=M-\overline{M}$
is the centered pre-adaptation \textsc{FlowEval} metric. The coefficient of interest is ($\beta_3$): it tests whether the change in success rate associated with additional demonstrations depends on the initial \textsc{FlowEval} value. We fit the model with ordinary least squares using task-clustered robust covariance and test ($H_0$:$\beta_3=0$) with a two-sided Wald test. 

Under deployment shift, we observe significant interactions for both Pi-05 and GR00T-N1.5 (Figure~\ref{fig:regression}). For Pi-05, both velocity error and cosine similarity are associated with how success rate changes across the 50, 150, and 500 demonstration conditions. GR00T-N1.5 shows the same relationship over the 50 and 150 demonstration conditions. 
Across both model families, \textsc{FlowEval} therefore provides information before adaptation about how performance changes as additional target-domain data is provided.

In contrast, we do not observe significant interactions in the control experiments with Pi-05-RoboCasa and GR00T-N1.6, where the deployment setting is already represented during pretraining (Appendix~\ref{appendix:sec:deployment-shift-control}). 
This contrast supports the interpretation that the interaction is conditioned on deployment shift. Additional controls further ruled out alternative explanations: Within a model, the interaction persists after accounting for success rate improvement rooms (Appendix~\ref{appendix:sec:headroom-control}); Across models, we did not find a shared task-difficulty pattern(Appendix~\ref{appendix:sec:task-control}) explaining the corelation.

Notably, the relationship between \textsc{FlowEval} scores and finetuned performance depends on the adaptation setting. When adaptation crosses action spaces, action-representation mismatch introduces an additional source of predictive disagreement beyond prediction error. Lower error therefore need not imply higher finetuned success. This phenomenon is observed for Pi-05 in Fig.\ref{fig:regression}. We also observe that additional demonstrations sometimes reduce performance on GR00T-N1.5, consistent with prior observations \citep{park2026behavioral}. Training configurations may affect the direction and shape of these curves (Appendix~\ref{appendix:sec:training-configs}). 
The significant interactions observed across both model families nevertheless suggest that \textsc{FlowEval}'s predictive relationship with performance--data interaction extends beyond a single model and training configuration.
\vspace{-0.4em}

\begin{figure*}[htbp]
    \centering
    \begin{minipage}{0.47\textwidth}
    \subcaption{Pi-05}
        \centering
        \includegraphics[width=\linewidth]{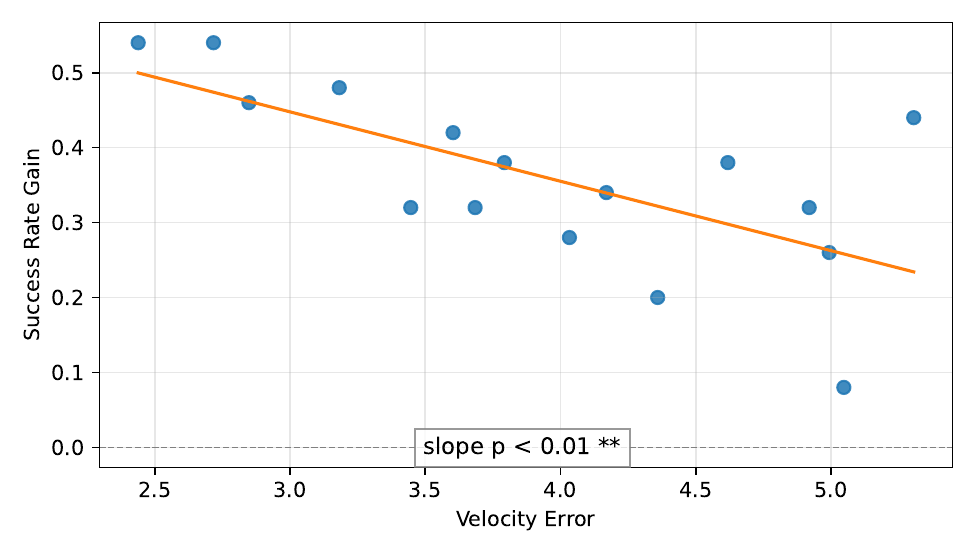}
    \end{minipage}
    \hfill
    \begin{minipage}{0.47\textwidth}
    \subcaption{GR00T-N1.5}
        \centering
        \includegraphics[width=\linewidth]{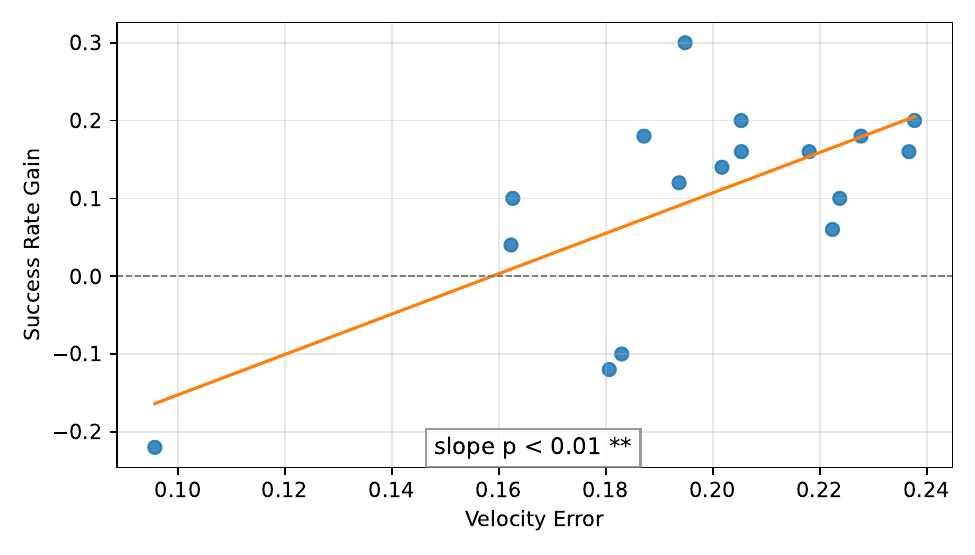}
    \end{minipage}

    \begin{minipage}{0.47\textwidth}
        \centering
        \includegraphics[width=\linewidth]{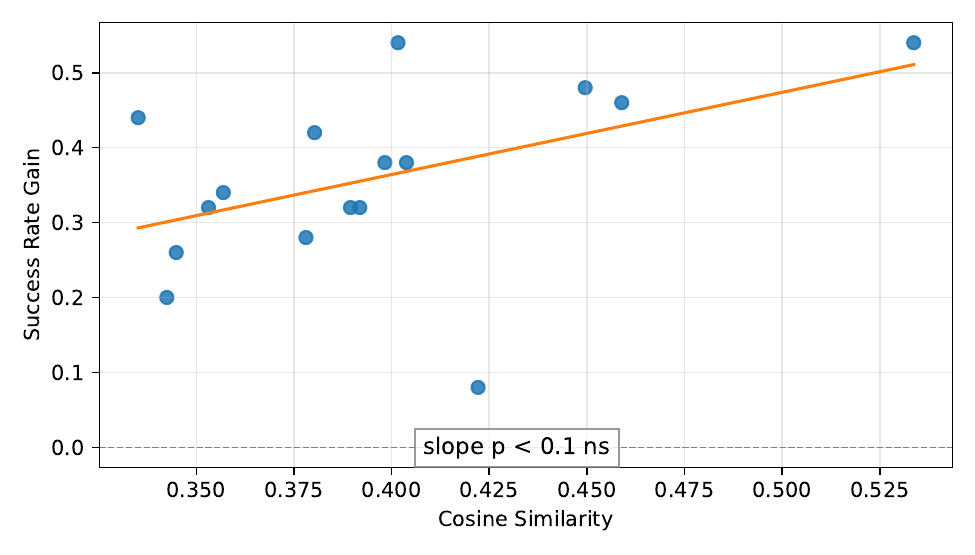}
    \end{minipage}
    \hfill
    \begin{minipage}{0.47\textwidth}
        \centering
        \includegraphics[width=\linewidth]{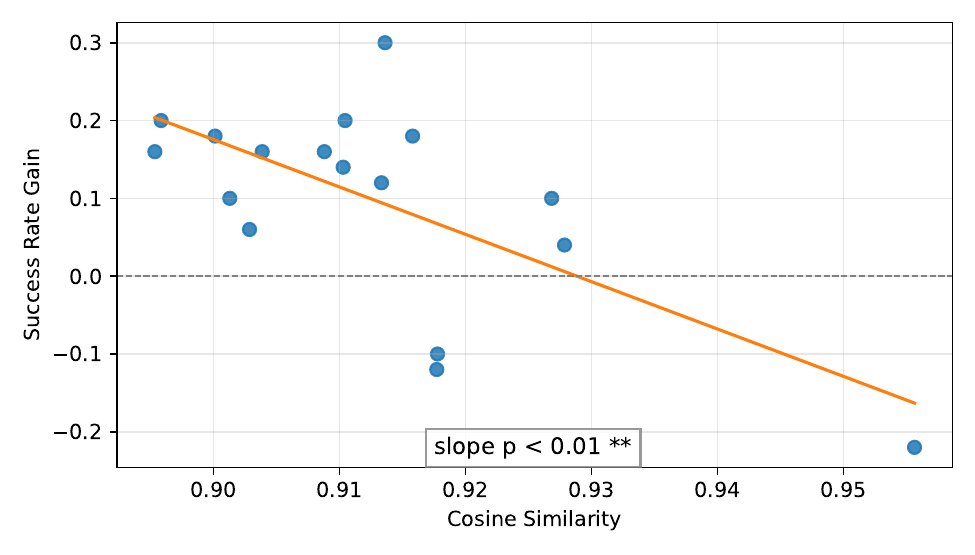}
    \end{minipage}
    \caption{Relation between performance gain and metric values. 500 - 50 demos success rate for Pi-05, 150 - 50 demos success rate for GR00T-N1.5.}
    \label{fig:performance-gain}
\end{figure*}

\vspace{-0.4em}
\section{Conclusion and Future Work}

We introduced \textsc{FlowEval}, a training-free diagnostic for probing how well a pretrained embodied foundation model fits a target deployment. Across multiple model families and deployment settings, \textsc{FlowEval} exhibits greater predictive agreement on deployment settings represented during pretraining and, under deployment shift, correlates with how performance scales as additional adaptation data is provided.

Future work should test whether these signals generalize across broader model and task families, and whether they can support calibrated predictions of the amount of adaptation data required to reach a target level of performance. More broadly, the same idea may extend from diagnosing \emph{how much adaptation remains} to diagnosing \emph{what kind of prior is missing}. Modern embodied models increasingly combine heterogeneous sources of prior knowledge---robot trajectories, human video, simulation, world-model pretraining, and large-scale vision-language representations---whose contributions may differ substantially across downstream behaviors. A natural question is whether model-based diagnostics can estimate how much a particular source of additional pretraining would improve a model's fit to a target deployment, before actually carrying out that pretraining.

Such an analysis would require controlled pretraining interventions across data sources, model families, and scales, and is therefore beyond the scope of an individual academic study. At sufficient scale, however, it could turn the composition of embodied pretraining into a measurable allocation problem: given a model and a family of target deployments, which additional source of experience most efficiently closes the remaining gap?

\clearpage
\acknowledgments{This research used resources provided through NSF ACCESS allocation CIS250679. We are grateful to Dr. Joseph Campbell for helping us obtain this allocation.
}


\bibliography{example}  

@misc{kim2024openvla,
      title={OpenVLA: An Open-Source Vision-Language-Action Model}, 
      author={Moo Jin Kim and Karl Pertsch and Siddharth Karamcheti and Ted Xiao and Ashwin Balakrishna and Suraj Nair and Rafael Rafailov and Ethan Foster and Grace Lam and Pannag Sanketi and Quan Vuong and Thomas Kollar and Benjamin Burchfiel and Russ Tedrake and Dorsa Sadigh and Sergey Levine and Percy Liang and Chelsea Finn},
      year={2024},
      eprint={2406.09246},
      archivePrefix={arXiv},
      primaryClass={cs.RO},
      url={https://arxiv.org/abs/2406.09246}, 
}

@misc{ye2026dreamzero,
      title={World Action Models are Zero-shot Policies}, 
      author={Seonghyeon Ye and Yunhao Ge and Kaiyuan Zheng and Shenyuan Gao and Sihyun Yu and George Kurian and Suneel Indupuru and You Liang Tan and Chuning Zhu and Jiannan Xiang and Ayaan Malik and Kyungmin Lee and William Liang and Nadun Ranawaka and Jiasheng Gu and Yinzhen Xu and Guanzhi Wang and Fengyuan Hu and Avnish Narayan and Johan Bjorck and Jing Wang and Gwanghyun Kim and Dantong Niu and Ruijie Zheng and Yuqi Xie and Jimmy Wu and Qi Wang and Ryan Julian and Danfei Xu and Yilun Du and Yevgen Chebotar and Scott Reed and Jan Kautz and Yuke Zhu and Linxi "Jim" Fan and Joel Jang},
      year={2026},
      eprint={2602.15922},
      archivePrefix={arXiv},
      primaryClass={cs.RO},
      url={https://arxiv.org/abs/2602.15922}, 
}

@misc{fang2026molmoact2,
      title={MolmoAct2: Action Reasoning Models for Real-world Deployment}, 
      author={Haoquan Fang and Jiafei Duan and Donovan Clay and Sam Wang and Shuo Liu and Weikai Huang and Xiang Fan and Wei-Chuan Tsai and Shirui Chen and Yi Ru Wang and Shanli Xing and Jaemin Cho and Jae Sung Park and Ainaz Eftekhar and Peter Sushko and Karen Farley and Angad Wadhwa and Cole Harrison and Winson Han and Ying-Chun Lee and Eli VanderBilt and Rose Hendrix and Suveen Ellawela and Lucas Ngoo and Joyce Chai and Zhongzheng Ren and Ali Farhadi and Dieter Fox and Ranjay Krishna},
      year={2026},
      eprint={2605.02881},
      archivePrefix={arXiv},
      primaryClass={cs.RO},
      url={https://arxiv.org/abs/2605.02881}, 
}

@inproceedings{gr00tn1_2025,
  archivePrefix = {arxiv},
  eprint     = {2503.14734},
  title      = {{GR00T} {N1}: An Open Foundation Model for Generalist Humanoid Robots},
  author     = {NVIDIA and Johan Bjorck and Fernando Castañeda, Nikita Cherniadev and Xingye Da and Runyu Ding and Linxi "Jim" Fan and Yu Fang and Dieter Fox and Fengyuan Hu and Spencer Huang and Joel Jang and Zhenyu Jiang and Jan Kautz and Kaushil Kundalia and Lawrence Lao and Zhiqi Li and Zongyu Lin and Kevin Lin and Guilin Liu and Edith Llontop and Loic Magne and Ajay Mandlekar and Avnish Narayan and Soroush Nasiriany and Scott Reed and You Liang Tan and Guanzhi Wang and Zu Wang and Jing Wang and Qi Wang and Jiannan Xiang and Yuqi Xie and Yinzhen Xu and Zhenjia Xu and Seonghyeon Ye and Zhiding Yu and Ao Zhang and Hao Zhang and Yizhou Zhao and Ruijie Zheng and Yuke Zhu},
  month      = {March},
  year       = {2025},
  booktitle  = {ArXiv Preprint},
}

@misc{khazatsky2025droid,
      title={DROID: A Large-Scale In-The-Wild Robot Manipulation Dataset}, 
      author={Alexander Khazatsky and Karl Pertsch and Suraj Nair and Ashwin Balakrishna and Sudeep Dasari and Siddharth Karamcheti and Soroush Nasiriany and Mohan Kumar Srirama and Lawrence Yunliang Chen and Kirsty Ellis and Peter David Fagan and Joey Hejna and Masha Itkina and Marion Lepert and Yecheng Jason Ma and Patrick Tree Miller and Jimmy Wu and Suneel Belkhale and Shivin Dass and Huy Ha and Arhan Jain and Abraham Lee and Youngwoon Lee and Marius Memmel and Sungjae Park and Ilija Radosavovic and Kaiyuan Wang and Albert Zhan and Kevin Black and Cheng Chi and Kyle Beltran Hatch and Shan Lin and Jingpei Lu and Jean Mercat and Abdul Rehman and Pannag R Sanketi and Archit Sharma and Cody Simpson and Quan Vuong and Homer Rich Walke and Blake Wulfe and Ted Xiao and Jonathan Heewon Yang and Arefeh Yavary and Tony Z. Zhao and Christopher Agia and Rohan Baijal and Mateo Guaman Castro and Daphne Chen and Qiuyu Chen and Trinity Chung and Jaimyn Drake and Ethan Paul Foster and Jensen Gao and Vitor Guizilini and David Antonio Herrera and Minho Heo and Kyle Hsu and Jiaheng Hu and Muhammad Zubair Irshad and Donovon Jackson and Charlotte Le and Yunshuang Li and Kevin Lin and Roy Lin and Zehan Ma and Abhiram Maddukuri and Suvir Mirchandani and Daniel Morton and Tony Nguyen and Abigail O'Neill and Rosario Scalise and Derick Seale and Victor Son and Stephen Tian and Emi Tran and Andrew E. Wang and Yilin Wu and Annie Xie and Jingyun Yang and Patrick Yin and Yunchu Zhang and Osbert Bastani and Glen Berseth and Jeannette Bohg and Ken Goldberg and Abhinav Gupta and Abhishek Gupta and Dinesh Jayaraman and Joseph J Lim and Jitendra Malik and Roberto Martín-Martín and Subramanian Ramamoorthy and Dorsa Sadigh and Shuran Song and Jiajun Wu and Michael C. Yip and Yuke Zhu and Thomas Kollar and Sergey Levine and Chelsea Finn},
      year={2025},
      eprint={2403.12945},
      archivePrefix={arXiv},
      primaryClass={cs.RO},
      url={https://arxiv.org/abs/2403.12945}, 
}

@misc{liu2023libero,
      title={LIBERO: Benchmarking Knowledge Transfer for Lifelong Robot Learning}, 
      author={Bo Liu and Yifeng Zhu and Chongkai Gao and Yihao Feng and Qiang Liu and Yuke Zhu and Peter Stone},
      year={2023},
      eprint={2306.03310},
      archivePrefix={arXiv},
      primaryClass={cs.AI},
      url={https://arxiv.org/abs/2306.03310}, 
}

@misc{physical_intelligence2025pi05,
      title={$\pi_{0.5}$: a Vision-Language-Action Model with Open-World Generalization}, 
      author={Physical Intelligence and Kevin Black and Noah Brown and James Darpinian and Karan Dhabalia and Danny Driess and Adnan Esmail and Michael Equi and Chelsea Finn and Niccolo Fusai and Manuel Y. Galliker and Dibya Ghosh and Lachy Groom and Karol Hausman and Brian Ichter and Szymon Jakubczak and Tim Jones and Liyiming Ke and Devin LeBlanc and Sergey Levine and Adrian Li-Bell and Mohith Mothukuri and Suraj Nair and Karl Pertsch and Allen Z. Ren and Lucy Xiaoyang Shi and Laura Smith and Jost Tobias Springenberg and Kyle Stachowicz and James Tanner and Quan Vuong and Homer Walke and Anna Walling and Haohuan Wang and Lili Yu and Ury Zhilinsky},
      year={2025},
      eprint={2504.16054},
      archivePrefix={arXiv},
      primaryClass={cs.LG},
      url={https://arxiv.org/abs/2504.16054}, 
}

@misc{park2026behavioral,
  author       = {Seohong Park},
  title        = {Behavioral Cloning Mystery},
  year         = {2026},
  month        = aug,
  howpublished = {\url{https://seohong.me/blog/behavioral-cloning-mystery/}},
  note         = {Blog post}
}

@article{skild2026s1,
  author = {Skild AI},
  title  = {Introducing S1: In-Context Learning for Robotics},
  journal = {Skild AI Blog},
  year   = {2026},
  month  = {August},
  url    = {https://skild.ai/blogs/s1},
}

@article{generalist2026gen15,
author = {Generalist Team},
title = {GEN-1.5: Embodied Foundation Models are One-Shot Learners},
journal = {Generalist AI Blog},
year = {2026},
note = {https://generalistai.com/blog/gen-1.5},
}

@misc{nasiriany2026robocasa365,
      title={RoboCasa365: A Large-Scale Simulation Framework for Training and Benchmarking Generalist Robots}, 
      author={Soroush Nasiriany and Sepehr Nasiriany and Abhiram Maddukuri and Yuke Zhu},
      year={2026},
      eprint={2603.04356},
      archivePrefix={arXiv},
      primaryClass={cs.RO},
      url={https://arxiv.org/abs/2603.04356}, 
}

@misc{deepmind2026geminirobotics2,
  author       = {{Google DeepMind}},
  title        = {Gemini Robotics 2},
  year         = {2026},
  howpublished = {\url{https://deepmind.google/models/gemini-robotics/vla/}},
  note         = {Model webpage}
}

@misc{geminiroboticsteam2025geminirobotics15,
      title={Gemini Robotics 1.5: Pushing the Frontier of Generalist Robots with Advanced Embodied Reasoning, Thinking, and Motion Transfer}, 
      author={Gemini Robotics Team and Abbas Abdolmaleki and Saminda Abeyruwan and Joshua Ainslie and Jean-Baptiste Alayrac and Montserrat Gonzalez Arenas and Ashwin Balakrishna and Nathan Batchelor and Alex Bewley and Jeff Bingham and Michael Bloesch and Konstantinos Bousmalis and Philemon Brakel and Anthony Brohan and Thomas Buschmann and Arunkumar Byravan and Serkan Cabi and Ken Caluwaerts and Federico Casarini and Christine Chan and Oscar Chang and London Chappellet-Volpini and Jose Enrique Chen and Xi Chen and Hao-Tien Lewis Chiang and Krzysztof Choromanski and Adrian Collister and David B. D'Ambrosio and Sudeep Dasari and Todor Davchev and Meet Kirankumar Dave and Coline Devin and Norman Di Palo and Tianli Ding and Carl Doersch and Adil Dostmohamed and Yilun Du and Debidatta Dwibedi and Sathish Thoppay Egambaram and Michael Elabd and Tom Erez and Xiaolin Fang and Claudio Fantacci and Cody Fong and Erik Frey and Chuyuan Fu and Ruiqi Gao and Marissa Giustina and Keerthana Gopalakrishnan and Laura Graesser and Oliver Groth and Agrim Gupta and Roland Hafner and Steven Hansen and Leonard Hasenclever and Sam Haves and Nicolas Heess and Brandon Hernaez and Alex Hofer and Jasmine Hsu and Lu Huang and Sandy H. Huang and Atil Iscen and Mithun George Jacob and Deepali Jain and Sally Jesmonth and Abhishek Jindal and Ryan Julian and Dmitry Kalashnikov and M. Emre Karagozler and Stefani Karp and Matija Kecman and J. Chase Kew and Donnie Kim and Frank Kim and Junkyung Kim and Thomas Kipf and Sean Kirmani and Ksenia Konyushkova and Li Yang Ku and Yuheng Kuang and Thomas Lampe and Antoine Laurens and Tuan Anh Le and Isabel Leal and Alex X. Lee and Tsang-Wei Edward Lee and Guy Lever and Jacky Liang and Li-Heng Lin and Fangchen Liu and Shangbang Long and Caden Lu and Sharath Maddineni and Anirudha Majumdar and Kevis-Kokitsi Maninis and Andrew Marmon and Sergio Martinez and Assaf Hurwitz Michaely and Niko Milonopoulos and Joss Moore and Robert Moreno and Michael Neunert and Francesco Nori and Joy Ortiz and Kenneth Oslund and Carolina Parada and Emilio Parisotto and Amaris Paryag and Acorn Pooley and Thomas Power and Alessio Quaglino and Haroon Qureshi and Rajkumar Vasudeva Raju and Helen Ran and Dushyant Rao and Kanishka Rao and Isaac Reid and David Rendleman and Krista Reymann and Miguel Rivas and Francesco Romano and Yulia Rubanova and Peter Pastor Sampedro and Pannag R Sanketi and Dhruv Shah and Mohit Sharma and Kathryn Shea and Mohit Shridhar and Charles Shu and Vikas Sindhwani and Sumeet Singh and Radu Soricut and Rachel Sterneck and Ian Storz and Razvan Surdulescu and Jie Tan and Jonathan Tompson and Saran Tunyasuvunakool and Jake Varley and Grace Vesom and Giulia Vezzani and Maria Bauza Villalonga and Oriol Vinyals and René Wagner and Ayzaan Wahid and Stefan Welker and Paul Wohlhart and Chengda Wu and Markus Wulfmeier and Fei Xia and Ted Xiao and Annie Xie and Jinyu Xie and Peng Xu and Sichun Xu and Ying Xu and Zhuo Xu and Jimmy Yan and Sherry Yang and Skye Yang and Yuxiang Yang and Hiu Hong Yu and Wenhao Yu and Wentao Yuan and Yuan Yuan and Jingwei Zhang and Tingnan Zhang and Zhiyuan Zhang and Allan Zhou and Guangyao Zhou and Yuxiang Zhou},
      year={2025},
      eprint={2510.03342},
      archivePrefix={arXiv},
      primaryClass={cs.RO},
      url={https://arxiv.org/abs/2510.03342}, 
}

@misc{figure2026helix02,
  author  = {Figure},
  title   = {Introducing Helix 02: Full-Body Autonomy},
  journal = {Figure AI Blog},
  date    = {2026-01-27},
  year = {2026}, 
  url     = {https://www.figure.ai/news/helix-02},
  urldate = {2026-09-13}
}

@article{genesis2026gene265, 
author = {Genesis AI Team}, 
title = {GENE-26.5: Advancing Robotic Manipulation to Human Level},
journal = {Genesis AI Blog}, 
month = {May}, 
year = {2026}, 
url = {https://genesis.ai/blog/gene-26-5-advancing-robotic-manipulation-to-human-level},}

@inproceedings{octo_2023,
    title={Octo: An Open-Source Generalist Robot Policy},
    author = {{Octo Model Team} and Dibya Ghosh and Homer Walke and Karl Pertsch and Kevin Black and Oier Mees and Sudeep Dasari and Joey Hejna and Charles Xu and Jianlan Luo and Tobias Kreiman and {You Liang} Tan and Lawrence Yunliang Chen and Pannag Sanketi and Quan Vuong and Ted Xiao and Dorsa Sadigh and Chelsea Finn and Sergey Levine},
    booktitle = {Proceedings of Robotics: Science and Systems},
    address  = {Delft, Netherlands},
    year = {2024},
}

@misc{kim2025finetuningvisionlanguageactionmodelsoptimizing,
      title={Fine-Tuning Vision-Language-Action Models: Optimizing Speed and Success}, 
      author={Moo Jin Kim and Chelsea Finn and Percy Liang},
      year={2025},
      eprint={2502.19645},
      archivePrefix={arXiv},
      primaryClass={cs.RO},
      url={https://arxiv.org/abs/2502.19645}, 
}

@misc{kaplan2020scalinglawsneurallanguage,
      title={Scaling Laws for Neural Language Models}, 
      author={Jared Kaplan and Sam McCandlish and Tom Henighan and Tom B. Brown and Benjamin Chess and Rewon Child and Scott Gray and Alec Radford and Jeffrey Wu and Dario Amodei},
      year={2020},
      eprint={2001.08361},
      archivePrefix={arXiv},
      primaryClass={cs.LG},
      url={https://arxiv.org/abs/2001.08361}, 
}

@misc{hoffmann2022trainingcomputeoptimallargelanguage,
      title={Training Compute-Optimal Large Language Models}, 
      author={Jordan Hoffmann and Sebastian Borgeaud and Arthur Mensch and Elena Buchatskaya and Trevor Cai and Eliza Rutherford and Diego de Las Casas and Lisa Anne Hendricks and Johannes Welbl and Aidan Clark and Tom Hennigan and Eric Noland and Katie Millican and George van den Driessche and Bogdan Damoc and Aurelia Guy and Simon Osindero and Karen Simonyan and Erich Elsen and Jack W. Rae and Oriol Vinyals and Laurent Sifre},
      year={2022},
      eprint={2203.15556},
      archivePrefix={arXiv},
      primaryClass={cs.CL},
      url={https://arxiv.org/abs/2203.15556}, 
}

@misc{zhao2023learningfinegrainedbimanualmanipulation,
      title={Learning Fine-Grained Bimanual Manipulation with Low-Cost Hardware}, 
      author={Tony Z. Zhao and Vikash Kumar and Sergey Levine and Chelsea Finn},
      year={2023},
      eprint={2304.13705},
      archivePrefix={arXiv},
      primaryClass={cs.RO},
      url={https://arxiv.org/abs/2304.13705}, 
}

@misc{zhang2017understandingdeeplearningrequires,
      title={Understanding deep learning requires rethinking generalization}, 
      author={Chiyuan Zhang and Samy Bengio and Moritz Hardt and Benjamin Recht and Oriol Vinyals},
      year={2017},
      eprint={1611.03530},
      archivePrefix={arXiv},
      primaryClass={cs.LG},
      url={https://arxiv.org/abs/1611.03530}, 
}

@misc{guo2017calibrationmodernneuralnetworks,
      title={On Calibration of Modern Neural Networks}, 
      author={Chuan Guo and Geoff Pleiss and Yu Sun and Kilian Q. Weinberger},
      year={2017},
      eprint={1706.04599},
      archivePrefix={arXiv},
      primaryClass={cs.LG},
      url={https://arxiv.org/abs/1706.04599}, 
}

@misc{li2024evaluatingrealworldrobotmanipulation,
      title={Evaluating Real-World Robot Manipulation Policies in Simulation}, 
      author={Xuanlin Li and Kyle Hsu and Jiayuan Gu and Karl Pertsch and Oier Mees and Homer Rich Walke and Chuyuan Fu and Ishikaa Lunawat and Isabel Sieh and Sean Kirmani and Sergey Levine and Jiajun Wu and Chelsea Finn and Hao Su and Quan Vuong and Ted Xiao},
      year={2024},
      eprint={2405.05941},
      archivePrefix={arXiv},
      primaryClass={cs.RO},
      url={https://arxiv.org/abs/2405.05941}, 
}

@misc{zhou2025autoevalautonomousevaluationgeneralist,
      title={AutoEval: Autonomous Evaluation of Generalist Robot Manipulation Policies in the Real World}, 
      author={Zhiyuan Zhou and Pranav Atreya and You Liang Tan and Karl Pertsch and Sergey Levine},
      year={2025},
      eprint={2503.24278},
      archivePrefix={arXiv},
      primaryClass={cs.RO},
      url={https://arxiv.org/abs/2503.24278}, 
}

@misc{han2026conditionalflowmatchingreplace,
      title={When Can Conditional Flow Matching Replace Pointwise Negative Log-Likelihood?}, 
      author={Yansen Han and Hongxin Sun and Tao Lin},
      year={2026},
      eprint={2608.28010},
      archivePrefix={arXiv},
      primaryClass={cs.LG},
      url={https://arxiv.org/abs/2608.28010}, 
}

@article{Wenger2025anomalously,
  author  = {Wenger, Ingeborg and Ebel, Henrik and Eberhard, Peter},
  title   = {Anomalously acting agents: the deployment problem},
  journal = {Multibody System Dynamics},
  year    = {2025},
  volume  = {64},
  number  = {3},
  pages   = {465--483},
  doi     = {10.1007/s11044-024-09993-1},
  url     = {https://doi.org/10.1007/s11044-024-09993-1},
  issn    = {1573-272X}
}

@misc{nalisnick2019deepgenerativemodelsknow,
      title={Do Deep Generative Models Know What They Don't Know?}, 
      author={Eric Nalisnick and Akihiro Matsukawa and Yee Whye Teh and Dilan Gorur and Balaji Lakshminarayanan},
      year={2019},
      eprint={1810.09136},
      archivePrefix={arXiv},
      primaryClass={stat.ML},
      url={https://arxiv.org/abs/1810.09136}, 
}

@inproceedings{NEURIPS2023_zhang_constrainedpolicy,
 author = {Zhang, Jing and Zhang, Chi and Wang, Wenjia and Jing, Bingyi},
 booktitle = {Advances in Neural Information Processing Systems},
 doi = {10.52202/075280-0245},
 editor = {A. Oh and T. Naumann and A. Globerson and K. Saenko and M. Hardt and S. Levine},
 pages = {5616--5630},
 publisher = {Curran Associates, Inc.},
 title = {Constrained Policy Optimization with Explicit Behavior Density For Offline Reinforcement Learning},
 url = {https://proceedings.neurips.cc/paper_files/paper/2023/file/11e1900e680f5fe1893a8e27362dbe2c-Paper-Conference.pdf},
 volume = {36},
 year = {2023}
}

@misc{chae2026flowactorcriticofflinereinforcement,
      title={Flow Actor-Critic for Offline Reinforcement Learning}, 
      author={Jongseong Chae and Jongeui Park and Yongjae Shin and Gyeongmin Kim and Seungyul Han and Youngchul Sung},
      year={2026},
      eprint={2602.18015},
      archivePrefix={arXiv},
      primaryClass={cs.LG},
      url={https://arxiv.org/abs/2602.18015}, 
}

@misc{robocurve2026groot,
  author       = {{Robocurve}},
  title        = {{GR00T N1.7: SO-101 fine-tune on MolmoAct2 community data}},
  year         = {2026},
  howpublished = {Hugging Face},
  url          = {https://huggingface.co/robocurve/gr00t-n1.7-so101-molmoact2},
  note         = {Accessed: 2026-09-15}
}

@misc{zha2026laplanguageactionpretrainingenables,
      title={LAP: Language-Action Pre-Training Enables Zero-shot Cross-Embodiment Transfer}, 
      author={Lihan Zha and Asher J. Hancock and Mingtong Zhang and Tenny Yin and Yixuan Huang and Dhruv Shah and Allen Z. Ren and Anirudha Majumdar},
      year={2026},
      eprint={2602.10556},
      archivePrefix={arXiv},
      primaryClass={cs.RO},
      url={https://arxiv.org/abs/2602.10556}, 
}

@misc{durkan2019neuralsplineflows,
      title={Neural Spline Flows}, 
      author={Conor Durkan and Artur Bekasov and Iain Murray and George Papamakarios},
      year={2019},
      eprint={1906.04032},
      archivePrefix={arXiv},
      primaryClass={stat.ML},
      url={https://arxiv.org/abs/1906.04032}, 
}

@misc{ross2011reductionimitationlearningstructured,
      title={A Reduction of Imitation Learning and Structured Prediction to No-Regret Online Learning}, 
      author={Stephane Ross and Geoffrey J. Gordon and J. Andrew Bagnell},
      year={2011},
      eprint={1011.0686},
      archivePrefix={arXiv},
      primaryClass={cs.LG},
      url={https://arxiv.org/abs/1011.0686}, 
}

@article{Papamakarios2021normalizing,
  author  = {George Papamakarios and Eric Nalisnick and Danilo Jimenez Rezende and Shakir Mohamed and Balaji Lakshminarayanan},
  title   = {Normalizing Flows for Probabilistic Modeling and Inference},
  journal = {Journal of Machine Learning Research},
  year    = {2021},
  volume  = {22},
  number  = {57},
  pages   = {1--64},
  url     = {http://jmlr.org/papers/v22/19-1028.html}
}

@article{rewardai2026om1,
  author = {Reward AI Team},
  title = {OM-1: Frontier Robot Intelligence, Learned Firsthand from Humans},
  journal = {Reward AI Blog},
  year = {2026},
  note = {https://rewardai.com/blog/OM-1/}
}

@misc{xue2025demogensyntheticdemonstrationgeneration,
      title={DemoGen: Synthetic Demonstration Generation for Data-Efficient Visuomotor Policy Learning}, 
      author={Zhengrong Xue and Shuying Deng and Zhenyang Chen and Yixuan Wang and Zhecheng Yuan and Huazhe Xu},
      year={2025},
      eprint={2502.16932},
      archivePrefix={arXiv},
      primaryClass={cs.RO},
      url={https://arxiv.org/abs/2502.16932}, 
}

@misc{mandlekar2023mimicgendatagenerationscalable,
      title={MimicGen: A Data Generation System for Scalable Robot Learning using Human Demonstrations}, 
      author={Ajay Mandlekar and Soroush Nasiriany and Bowen Wen and Iretiayo Akinola and Yashraj Narang and Linxi Fan and Yuke Zhu and Dieter Fox},
      year={2023},
      eprint={2310.17596},
      archivePrefix={arXiv},
      primaryClass={cs.RO},
      url={https://arxiv.org/abs/2310.17596}, 
}

@misc{bharadhwaj2023roboagentgeneralizationefficiencyrobot,
      title={RoboAgent: Generalization and Efficiency in Robot Manipulation via Semantic Augmentations and Action Chunking}, 
      author={Homanga Bharadhwaj and Jay Vakil and Mohit Sharma and Abhinav Gupta and Shubham Tulsiani and Vikash Kumar},
      year={2023},
      eprint={2309.01918},
      archivePrefix={arXiv},
      primaryClass={cs.RO},
      url={https://arxiv.org/abs/2309.01918}, 
}

@misc{brohan2023rt2,
      title={RT-2: Vision-Language-Action Models Transfer Web Knowledge to Robotic Control}, 
      author={Anthony Brohan and Noah Brown and Justice Carbajal and Yevgen Chebotar and Xi Chen and Krzysztof Choromanski and Tianli Ding and Danny Driess and Avinava Dubey and Chelsea Finn and Pete Florence and Chuyuan Fu and Montse Gonzalez Arenas and Keerthana Gopalakrishnan and Kehang Han and Karol Hausman and Alexander Herzog and Jasmine Hsu and Brian Ichter and Alex Irpan and Nikhil Joshi and Ryan Julian and Dmitry Kalashnikov and Yuheng Kuang and Isabel Leal and Lisa Lee and Tsang-Wei Edward Lee and Sergey Levine and Yao Lu and Henryk Michalewski and Igor Mordatch and Karl Pertsch and Kanishka Rao and Krista Reymann and Michael Ryoo and Grecia Salazar and Pannag Sanketi and Pierre Sermanet and Jaspiar Singh and Anikait Singh and Radu Soricut and Huong Tran and Vincent Vanhoucke and Quan Vuong and Ayzaan Wahid and Stefan Welker and Paul Wohlhart and Jialin Wu and Fei Xia and Ted Xiao and Peng Xu and Sichun Xu and Tianhe Yu and Brianna Zitkovich},
      year={2023},
      eprint={2307.15818},
      archivePrefix={arXiv},
      primaryClass={cs.RO},
      url={https://arxiv.org/abs/2307.15818}, 
}

@misc{nasiriany2026robocasa365largescalesimulationframework,
      title={RoboCasa365: A Large-Scale Simulation Framework for Training and Benchmarking Generalist Robots}, 
      author={Soroush Nasiriany and Sepehr Nasiriany and Abhiram Maddukuri and Yuke Zhu},
      year={2026},
      eprint={2603.04356},
      archivePrefix={arXiv},
      primaryClass={cs.RO},
      url={https://arxiv.org/abs/2603.04356}, 
}

@misc{liu2023liberobenchmarkingknowledgetransfer,
      title={LIBERO: Benchmarking Knowledge Transfer for Lifelong Robot Learning}, 
      author={Bo Liu and Yifeng Zhu and Chongkai Gao and Yihao Feng and Qiang Liu and Yuke Zhu and Peter Stone},
      year={2023},
      eprint={2306.03310},
      archivePrefix={arXiv},
      primaryClass={cs.AI},
      url={https://arxiv.org/abs/2306.03310}, 
}

@misc{mees2022calvinbenchmarklanguageconditionedpolicy,
      title={CALVIN: A Benchmark for Language-Conditioned Policy Learning for Long-Horizon Robot Manipulation Tasks}, 
      author={Oier Mees and Lukas Hermann and Erick Rosete-Beas and Wolfram Burgard},
      year={2022},
      eprint={2112.03227},
      archivePrefix={arXiv},
      primaryClass={cs.RO},
      url={https://arxiv.org/abs/2112.03227}, 
}

@article{li2024behavior1k,
    title   = {BEHAVIOR-1K: A Human-Centered, Embodied AI Benchmark with 1,000 Everyday Activities and Realistic Simulation},
    author  = {Chengshu Li and Ruohan Zhang and Josiah Wong and Cem Gokmen and Sanjana Srivastava and Roberto Martín-Martín and Chen Wang and Gabrael Levine and Wensi Ai and Benjamin Martinez and Hang Yin and Michael Lingelbach and Minjune Hwang and Ayano Hiranaka and Sujay Garlanka and Arman Aydin and Sharon Lee and Jiankai Sun and Mona Anvari and Manasi Sharma and Dhruva Bansal and Samuel Hunter and Kyu-Young Kim and Alan Lou and Caleb R Matthews and Ivan Villa-Renteria and Jerry Huayang Tang and Claire Tang and Fei Xia and Yunzhu Li and Silvio Savarese and Hyowon Gweon and C. Karen Liu and Jiajun Wu and Li Fei-Fei},
    journal = {arXiv preprint arXiv:2403.09227},
    year    = {2024}
}

@misc{james2019rlbenchrobotlearningbenchmark,
      title={RLBench: The Robot Learning Benchmark \& Learning Environment}, 
      author={Stephen James and Zicong Ma and David Rovick Arrojo and Andrew J. Davison},
      year={2019},
      eprint={1909.12271},
      archivePrefix={arXiv},
      primaryClass={cs.RO},
      url={https://arxiv.org/abs/1909.12271}, 
}

@misc{zhang2024vlabenchlargescalebenchmarklanguageconditioned,
      title={VLABench: A Large-Scale Benchmark for Language-Conditioned Robotics Manipulation with Long-Horizon Reasoning Tasks}, 
      author={Shiduo Zhang and Zhe Xu and Peiju Liu and Xiaopeng Yu and Yuan Li and Qinghui Gao and Zhaoye Fei and Zhangyue Yin and Zuxuan Wu and Yu-Gang Jiang and Xipeng Qiu},
      year={2024},
      eprint={2412.18194},
      archivePrefix={arXiv},
      primaryClass={cs.RO},
      url={https://arxiv.org/abs/2412.18194}, 
}

@misc{embodimentcollaboration2025openxembodimentroboticlearning,
      title={Open X-Embodiment: Robotic Learning Datasets and RT-X Models}, 
      author={Embodiment Collaboration and Abby O'Neill and Abdul Rehman and Abhinav Gupta and Abhiram Maddukuri and Abhishek Gupta and Abhishek Padalkar and Abraham Lee and Acorn Pooley and Agrim Gupta and Ajay Mandlekar and Ajinkya Jain and Albert Tung and Alex Bewley and Alex Herzog and Alex Irpan and Alexander Khazatsky and Anant Rai and Anchit Gupta and Andrew Wang and Andrey Kolobov and Anikait Singh and Animesh Garg and Aniruddha Kembhavi and Annie Xie and Anthony Brohan and Antonin Raffin and Archit Sharma and Arefeh Yavary and Arhan Jain and Ashwin Balakrishna and Ayzaan Wahid and Ben Burgess-Limerick and Beomjoon Kim and Bernhard Schölkopf and Blake Wulfe and Brian Ichter and Cewu Lu and Charles Xu and Charlotte Le and Chelsea Finn and Chen Wang and Chenfeng Xu and Cheng Chi and Chenguang Huang and Christine Chan and Christopher Agia and Chuer Pan and Chuyuan Fu and Coline Devin and Danfei Xu and Daniel Morton and Danny Driess and Daphne Chen and Deepak Pathak and Dhruv Shah and Dieter Büchler and Dinesh Jayaraman and Dmitry Kalashnikov and Dorsa Sadigh and Edward Johns and Ethan Foster and Fangchen Liu and Federico Ceola and Fei Xia and Feiyu Zhao and Felipe Vieira Frujeri and Freek Stulp and Gaoyue Zhou and Gaurav S. Sukhatme and Gautam Salhotra and Ge Yan and Gilbert Feng and Giulio Schiavi and Glen Berseth and Gregory Kahn and Guangwen Yang and Guanzhi Wang and Hao Su and Hao-Shu Fang and Haochen Shi and Henghui Bao and Heni Ben Amor and Henrik I Christensen and Hiroki Furuta and Homanga Bharadhwaj and Homer Walke and Hongjie Fang and Huy Ha and Igor Mordatch and Ilija Radosavovic and Isabel Leal and Jacky Liang and Jad Abou-Chakra and Jaehyung Kim and Jaimyn Drake and Jan Peters and Jan Schneider and Jasmine Hsu and Jay Vakil and Jeannette Bohg and Jeffrey Bingham and Jeffrey Wu and Jensen Gao and Jiaheng Hu and Jiajun Wu and Jialin Wu and Jiankai Sun and Jianlan Luo and Jiayuan Gu and Jie Tan and Jihoon Oh and Jimmy Wu and Jingpei Lu and Jingyun Yang and Jitendra Malik and João Silvério and Joey Hejna and Jonathan Booher and Jonathan Tompson and Jonathan Yang and Jordi Salvador and Joseph J. Lim and Junhyek Han and Kaiyuan Wang and Kanishka Rao and Karl Pertsch and Karol Hausman and Keegan Go and Keerthana Gopalakrishnan and Ken Goldberg and Kendra Byrne and Kenneth Oslund and Kento Kawaharazuka and Kevin Black and Kevin Lin and Kevin Zhang and Kiana Ehsani and Kiran Lekkala and Kirsty Ellis and Krishan Rana and Krishnan Srinivasan and Kuan Fang and Kunal Pratap Singh and Kuo-Hao Zeng and Kyle Hatch and Kyle Hsu and Laurent Itti and Lawrence Yunliang Chen and Lerrel Pinto and Li Fei-Fei and Liam Tan and Linxi "Jim" Fan and Lionel Ott and Lisa Lee and Luca Weihs and Magnum Chen and Marion Lepert and Marius Memmel and Masayoshi Tomizuka and Masha Itkina and Mateo Guaman Castro and Max Spero and Maximilian Du and Michael Ahn and Michael C. Yip and Mingtong Zhang and Mingyu Ding and Minho Heo and Mohan Kumar Srirama and Mohit Sharma and Moo Jin Kim and Muhammad Zubair Irshad and Naoaki Kanazawa and Nicklas Hansen and Nicolas Heess and Nikhil J Joshi and Niko Suenderhauf and Ning Liu and Norman Di Palo and Nur Muhammad Mahi Shafiullah and Oier Mees and Oliver Kroemer and Osbert Bastani and Pannag R Sanketi and Patrick "Tree" Miller and Patrick Yin and Paul Wohlhart and Peng Xu and Peter David Fagan and Peter Mitrano and Pierre Sermanet and Pieter Abbeel and Priya Sundaresan and Qiuyu Chen and Quan Vuong and Rafael Rafailov and Ran Tian and Ria Doshi and Roberto Martín-Martín and Rohan Baijal and Rosario Scalise and Rose Hendrix and Roy Lin and Runjia Qian and Ruohan Zhang and Russell Mendonca and Rutav Shah and Ryan Hoque and Ryan Julian and Samuel Bustamante and Sean Kirmani and Sergey Levine and Shan Lin and Sherry Moore and Shikhar Bahl and Shivin Dass and Shubham Sonawani and Shubham Tulsiani and Shuran Song and Sichun Xu and Siddhant Haldar and Siddharth Karamcheti and Simeon Adebola and Simon Guist and Soroush Nasiriany and Stefan Schaal and Stefan Welker and Stephen Tian and Subramanian Ramamoorthy and Sudeep Dasari and Suneel Belkhale and Sungjae Park and Suraj Nair and Suvir Mirchandani and Takayuki Osa and Tanmay Gupta and Tatsuya Harada and Tatsuya Matsushima and Ted Xiao and Thomas Kollar and Tianhe Yu and Tianli Ding and Todor Davchev and Tony Z. Zhao and Travis Armstrong and Trevor Darrell and Trinity Chung and Vidhi Jain and Vikash Kumar and Vincent Vanhoucke and Vitor Guizilini and Wei Zhan and Wenxuan Zhou and Wolfram Burgard and Xi Chen and Xiangyu Chen and Xiaolong Wang and Xinghao Zhu and Xinyang Geng and Xiyuan Liu and Xu Liangwei and Xuanlin Li and Yansong Pang and Yao Lu and Yecheng Jason Ma and Yejin Kim and Yevgen Chebotar and Yifan Zhou and Yifeng Zhu and Yilin Wu and Ying Xu and Yixuan Wang and Yonatan Bisk and Yongqiang Dou and Yoonyoung Cho and Youngwoon Lee and Yuchen Cui and Yue Cao and Yueh-Hua Wu and Yujin Tang and Yuke Zhu and Yunchu Zhang and Yunfan Jiang and Yunshuang Li and Yunzhu Li and Yusuke Iwasawa and Yutaka Matsuo and Zehan Ma and Zhuo Xu and Zichen Jeff Cui and Zichen Zhang and Zipeng Fu and Zipeng Lin},
      year={2025},
      eprint={2310.08864},
      archivePrefix={arXiv},
      primaryClass={cs.RO},
      url={https://arxiv.org/abs/2310.08864}, 
}

@misc{gururangan2020dontstoppretrainingadapt,
      title={Don't Stop Pretraining: Adapt Language Models to Domains and Tasks}, 
      author={Suchin Gururangan and Ana Marasović and Swabha Swayamdipta and Kyle Lo and Iz Beltagy and Doug Downey and Noah A. Smith},
      year={2020},
      eprint={2004.10964},
      archivePrefix={arXiv},
      primaryClass={cs.CL},
      url={https://arxiv.org/abs/2004.10964}, 
}

@inproceedings{moorelewis2010intelligent,
    title = "Intelligent Selection of Language Model Training Data",
    author = "Moore, Robert C.  and
      Lewis, William",
    editor = "Haji{\v{c}}, Jan  and
      Carberry, Sandra  and
      Clark, Stephen  and
      Nivre, Joakim",
    booktitle = "Proceedings of the {ACL} 2010 Conference Short Papers",
    month = jul,
    year = "2010",
    address = "Uppsala, Sweden",
    publisher = "Association for Computational Linguistics",
    url = "https://aclanthology.org/P10-2041/",
    pages = "220--224"
}

@inproceedings{DoReMi2023,
 author = {Xie, Sang Michael and Pham, Hieu and Dong, Xuanyi and Du, Nan and Liu, Hanxiao and Lu, Yifeng and Liang, Percy S and Le, Quoc V and Ma, Tengyu and Yu, Adams Wei},
 booktitle = {Advances in Neural Information Processing Systems},
 doi = {10.52202/075280-3059},
 editor = {A. Oh and T. Naumann and A. Globerson and K. Saenko and M. Hardt and S. Levine},
 pages = {69798--69818},
 publisher = {Curran Associates, Inc.},
 title = {DoReMi: Optimizing Data Mixtures Speeds Up Language Model Pretraining},
 url = {https://proceedings.neurips.cc/paper_files/paper/2023/file/dcba6be91359358c2355cd920da3fcbd-Paper-Conference.pdf},
 volume = {36},
 year = {2023}
}

@misc{lipman2023flowmatchinggenerativemodeling,
      title={Flow Matching for Generative Modeling}, 
      author={Yaron Lipman and Ricky T. Q. Chen and Heli Ben-Hamu and Maximilian Nickel and Matt Le},
      year={2023},
      eprint={2210.02747},
      archivePrefix={arXiv},
      primaryClass={cs.LG},
      url={https://arxiv.org/abs/2210.02747}, 
}

@misc{dreamzero-blog,
  author       = {Ye, Seonghyeon and others},
  title        = {{DreamZero}: World Action Models are Zero-shot Policies},
  year         = {2026},
  url          = {https://dreamzero0.github.io/},
  note         = {Accessed: 2026-10-04}
}

@misc{gr00tn16-blog,
  author       = {{GEAR Team} and others},
  title        = {{GR00T N1.6}: An Improved Open Foundation Model
                  for Generalist Humanoid Robots},
  year         = {2025},
  month        = dec,
  howpublished = {NVIDIA Research},
  url          = {https://research.nvidia.com/labs/gear/gr00t-n1_6/},
  note         = {Published December 15, 2025. Accessed: 2026-10-04}
}

@misc{pi05-blog,
  author       = {Black, Kevin and others},
  title        = {{$\pi_{0.5}$}: A {VLA} with Open-World Generalization},
  year         = {2025},
  month        = apr,
  howpublished = {Physical Intelligence},
  url          = {https://www.pi.website/blog/pi05},
  note         = {Published April 22, 2025. Accessed: 2026-10-04}
}

\newpage
\appendix
\section{Appendix}

\subsection{Design Choice}

Here we further discuss the design choices of \textsc{FlowEval}, including why pure noise is chosen over noisy actions; the first diffusion step is chosen over the rest; and cosine similarity and velocity error are chosen over action MSE. 

\subsubsection{Velocity Error vs. Validation loss}
\label{appendix:why-pure-noise}

\begin{figure}[htbp]
    \centering
    \includegraphics[width=\linewidth]{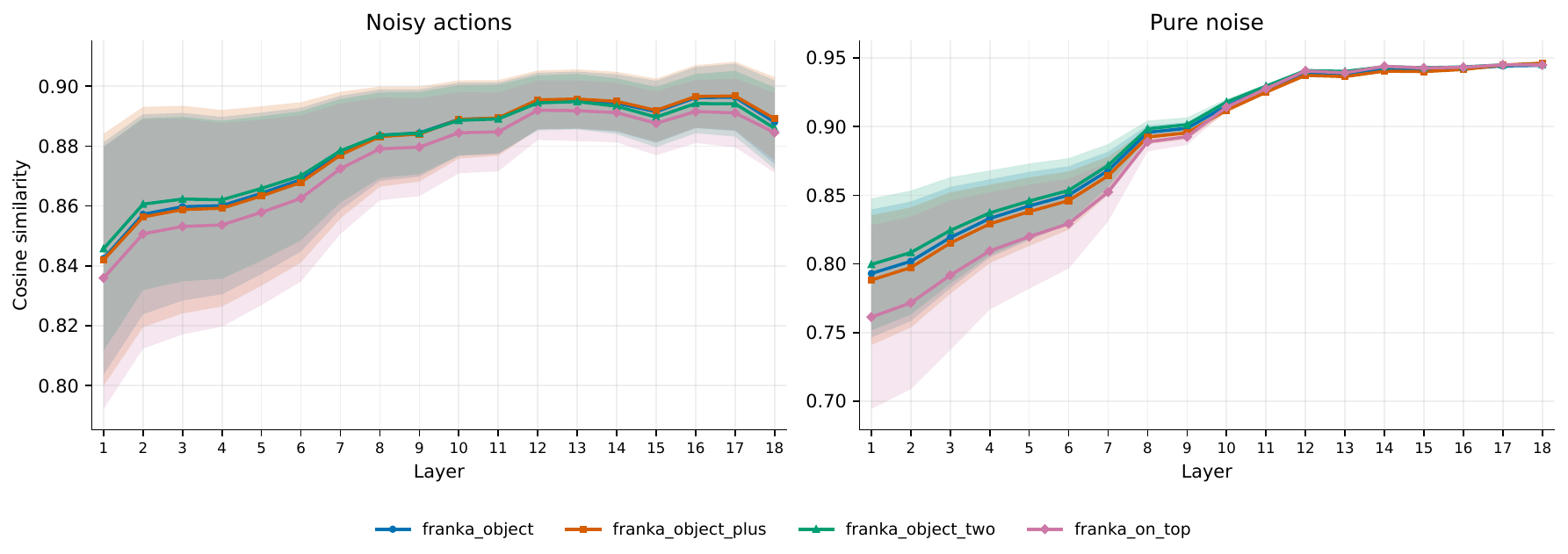}
    \caption{Cosine similarity per layer, evaluted on Pi-05. Left: noisy actions, right: pure noise.}
    \label{appendix:fig:cosine-per-layer}
\end{figure}

Validation loss is used to probe performance \citep{park2026behavioral, li2024evaluatingrealworldrobotmanipulation, zhou2025autoevalautonomousevaluationgeneralist}, yet flow matching introduces another complicacy in that the input noise can either be infused by ground truth action or not. We tested both conditions on four real-world Franka tasks, results in Fig.~\ref{appendix:fig:cosine-per-layer} and Table.~\ref{appendix:tab:noise_statistics}. Results show that pure noise yields notably more stable results than noisy actions, suggesting that it provides a more stable and reliable evaluation signal.

\begin{table}[htbp]
    \centering
    \caption{Statistics under noisy-action and pure-noise conditions on 4 collected Franka tasks.}
    \label{appendix:tab:noise_statistics}
    \begin{tabular}{llcc}
        \toprule
        \textbf{Task} & \textbf{Condition} & \textbf{Mean} & \textbf{Variance} \\
        \midrule
        \multirow{2}{*}{Franka Object}
            & Noisy actions & 3.121 & 3.390 \\
            & Pure noise    & 1.770 & 0.504 \\
        \midrule
        \multirow{2}{*}{Franka Object+}
            & Noisy actions & 3.089 & 3.286 \\
            & Pure noise    & 1.704 & 0.492 \\
        \midrule
        \multirow{2}{*}{Franka Object Two}
            & Noisy actions & 3.198 & 3.274 \\
            & Pure noise    & 1.775 & 0.513 \\
        \midrule
        \multirow{2}{*}{Franka On Top}
            & Noisy actions & 3.123 & 3.017 \\
            & Pure noise    & 1.762 & 0.540 \\
        \bottomrule
    \end{tabular}
\end{table}

\begin{table}[b]
\centering
\caption{Distinguishability score of Action MSE compared with \textsc{FlowEval Metrics}}
\label{appendix:fig:mse-gain}
\begin{tabular}{lccccc}
\toprule
Model & Action Cosine & Action VE & Action MSE & MSE / Cosine & MSE / VE \\
\midrule
$\pi_{0.5}$ & 0.816 & 1.07  & 1.10  & 1.35$\times$  & 1.03$\times$ \\
DreamZero    & 0.498 & 0.267 & 0.179 & 0.359$\times$ & 0.670$\times$ \\
\bottomrule
\end{tabular}
\end{table}

\subsubsection{Distinguishability Scores}
We include the full results used to generate Table \ref{tab:distinguishability} in Section \ref{sec:design_choices}. We ran 5 tasks on DreamZero-DROID, 16 diffusion steps, and 17 tasks on Pi-05-base, 10 diffusion steps. The tasks come from the 17 manipulation tasks within the Robocasa365 target dataset \citep{nasiriany2026robocasa365}. DreamZero-DROID requires significantly more computing resources than Pi-05, so we were unable to evaluate it on the full set.

Scores for every diffusion step are included in Table.~\ref{appendix:tab:pi05-full} and \ref{appendix:tab:dreamzero-full}, while advantage of choosing Action MSE over velocity error and cosine similarity is expressed by the division values in Table.~\ref{appendix:fig:mse-gain}. Choosing action MSE provides little gain for Pi-05, while causing a significant drop for DreamZero.


\begin{table*}[t]
\centering
\caption{DreamZero 16 diffusion steps}
\label{appendix:tab:dreamzero-full}
\resizebox{\textwidth}{!}{%
\begin{tabular}{lrrrrrrrr}
\hline
\textbf{Step} & \textbf{0} & \textbf{1} & \textbf{2} & \textbf{3} & \textbf{4} & \textbf{5} & \textbf{6} & \textbf{7} \\
\hline
Cosine (Action) & 0.498 & 0.455 & 0.398 & 0.398 & 0.398 & 0.398 & 0.323 & 0.323 \\
VE (Action)   & 0.267 & 0.244 & 0.205 & 0.205 & 0.205 & 0.205 & 0.144 & 0.144 \\
Cosine (Video)  & 1.17  & 0.799 & 0.799 & 0.799 & 0.799 & 0.799 & 1.07  & 1.07 \\
VE (Video)   & 0.956 & 0.735 & 0.723 & 0.723 & 0.723 & 0.723 & 1.14  & 1.14 \\
\hline
\textbf{Step} & \textbf{8} & \textbf{9} & \textbf{10} & \textbf{11} & \textbf{12} & \textbf{13} & \textbf{14} & \textbf{15} \\
\hline
Cosine (Action) & 0.323 & 0.323 & 0.231 & 0.231 & 0.231 & 0.224 & 0.234 & 0.231 \\
VE (Action)   & 0.144 & 0.144 & 0.130 & 0.130 & 0.130 & 0.158 & 0.170 & 0.180 \\
Cosine (Video)  & 1.07  & 1.07  & 0.897 & 0.897 & 0.897 & 0.845 & 0.846 & 0.850 \\
VE (Video)   & 1.14  & 1.14  & 0.906 & 0.906 & 0.906 & 0.830 & 0.830 & 0.854 \\
\hline
\end{tabular}%
}
\end{table*}

\begin{table*}[t]
\centering
\caption{Pi-05 10 diffusion steps}
\label{appendix:tab:pi05-full}
\begin{tabular}{lrrrrrrrrrr}
\hline
\textbf{Step} & \textbf{0} & \textbf{1} & \textbf{2} & \textbf{3} & \textbf{4} & \textbf{5} & \textbf{6} & \textbf{7} & \textbf{8} & \textbf{9}  \\
\hline
Cosine & 0.816 & 0.868 & 0.862 & 0.875 & 0.886 & 0.869 & 0.858 & 0.847 & 0.836 & 0.833 \\
VE & 1.07 & 1.10 & 1.10 & 1.10 & 1.10 & 1.10 & 1.10 & 1.10 & 1.10 & 1.09 \\
\hline
\end{tabular}
\end{table*}

\begin{figure}[b]
    \centering
    \begin{minipage}{0.32\textwidth}
    \subcaption{Action MSE}
        \centering
        \includegraphics[width=\linewidth]{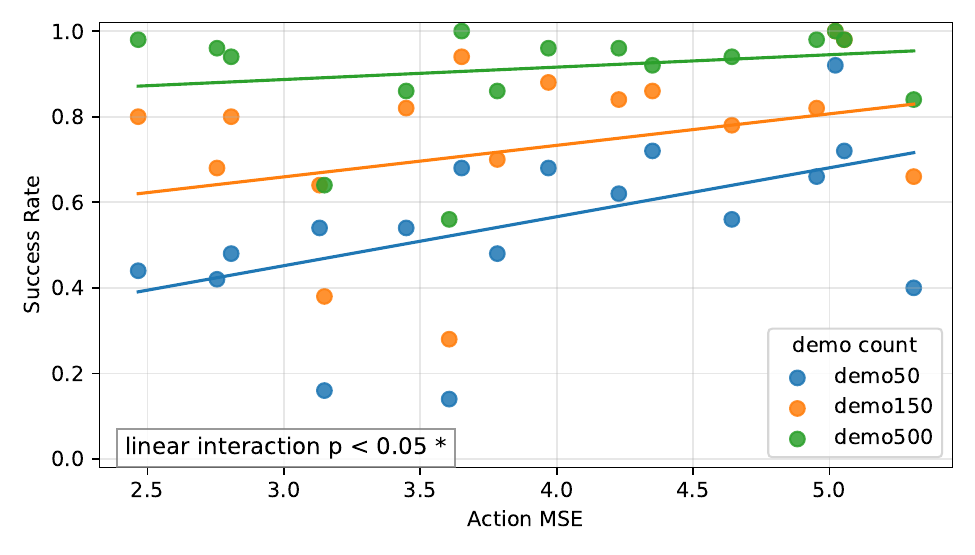}
    \end{minipage}
    \hfill
    \begin{minipage}{0.32\textwidth}
        \subcaption{Cosine Similarity}
        \centering
        \includegraphics[width=\linewidth]{figures/part1_regression/pi05-base_cosine_similarity.pdf}
    \end{minipage}
    \hfill
    \begin{minipage}{0.32\textwidth}
        \subcaption{Velocity Error}
        \centering
        \includegraphics[width=\linewidth]{figures/part1_regression/pi05-base_velocity_error.pdf}
    \end{minipage}
    \caption{Regression results in Sec.\ref{sec:regression}, recomputed with action MSE (left).}
    \label{appendix:fig:comparison}
\end{figure}

\subsubsection{Comparison with Action MSE}
\label{appendix:sec:action-mse}

Action MSE is a competitive baseline used to determine performance \citep{park2026behavioral}. It requires computing the whole diffusion process and thus can be up to 16 times more computationally inefficient, depending on the model configurations. However, faster computation does not necessarily translate to a better practical signal, raising the question of whether there is sufficient justification for designing \textsc{FlowEval} around the first diffusion step instead of action MSE.

We test from two perspectives (1) whether action MSE provide better distinguishability over a set of tasks compared with cosine similarity and velocity error, and (2) whether using action MSE instead of cosine similarity and velocity error changes our core results. We discovered that choosing action MSE provides little gain for Pi-05, while causing a significant drop for DreamZero.

However, distinguishability alone only lets us select a metric that's fairly separable (expressive), on top of that we still need practical signals showing the separation is actually reflective of the properties we care about. We therefore further evaluate whether the proposed \textsc{FlowEval} metrics reflect the downstream behavior we ultimately aim to predict. In particular, we show that \textsc{FlowEval} performs comparably to action MSE in correlating with how task success improves as additional adaptation data is provided (Figure.~\ref{appendix:fig:comparison}). Action MSE had a p value of .., cosine similarity.., velocity error..

\subsection{Limitation: Post transformed demos}
\label{appendix:counter-case}

A reasonable limitation of \textsc{FlowEval} is that it is designed for real human-teleoperated demonstrations, where 1) the evaluated action targets are the actions actually issued during data collection and 2) the demonstration was obtained with an actual rollout. Post-hoc transformed demonstrations do not necessarily satisfy this assumption and therefore may not be valid inputs to the metric. We do not attempt to enumerate every possible transformation. Instead, we present one concrete counterexample in which \textsc{FlowEval} does not apply reliably: demonstrations whose action targets are reconstructed after collection.

We collect the Franka demonstrations using a Cartesian-action controller. Given that Pi-05, MolmoAct2, GR00T-N1.6 and DreamZero pretrained real world Franka data in action spaces of delta joint-position, absolute joint-position, relative joint-position and relative joint-position respectively (Table.~\ref{appendix:tab:action_channels}), to evaluate these checkpoints within the same action representation they're pretrained to predict, we reconstruct joint-space action targets from the Cartesian demonstrations after collection.
This reconstruction creates a qualitatively different target from the original demonstration.

\begin{figure}[htbp]
    \centering
    \begin{minipage}{0.49\textwidth}
    \subcaption{Pi-05, Delta Joint-Position}
        \centering
        \includegraphics[width=\linewidth]{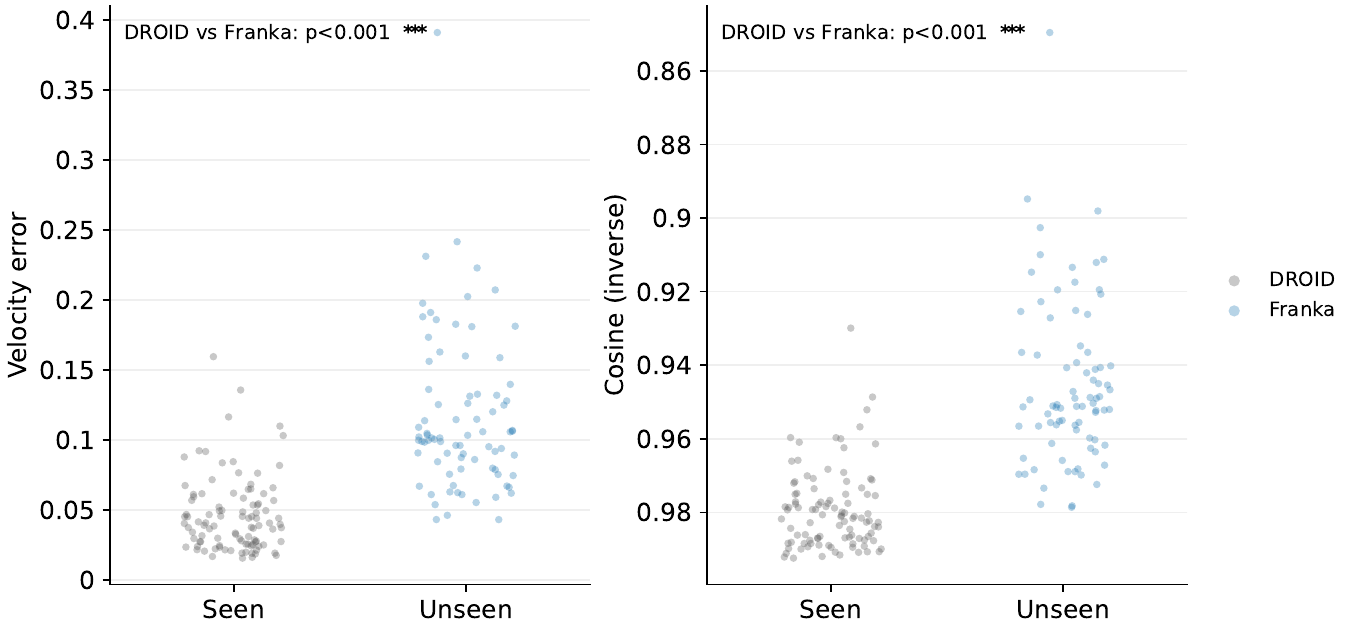}
    \end{minipage}
    \hfill
    \begin{minipage}{0.49\textwidth}
    \subcaption{MolmoAct2, Absolute Joint-Position}
        \centering
        \includegraphics[width=\linewidth]{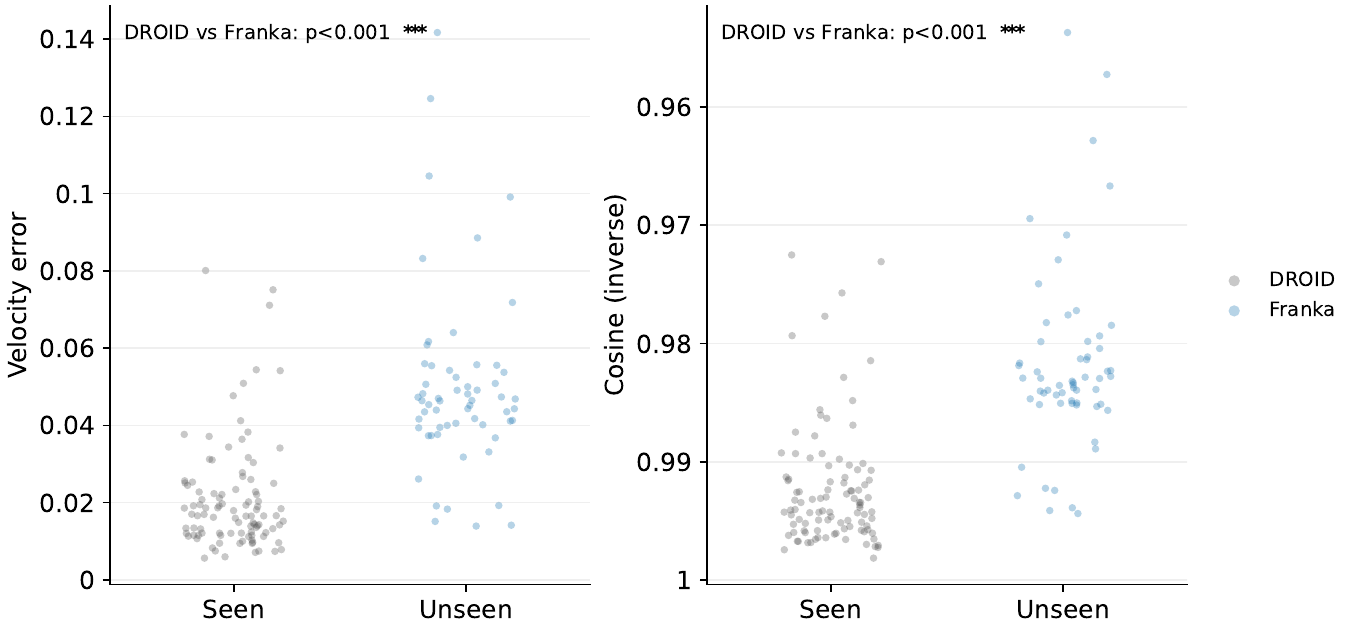}
    \end{minipage}

    \begin{minipage}{0.49\textwidth}
    \subcaption{GR00T-N1.6, Relative Joint-Position}
        \centering
        \includegraphics[width=\linewidth]{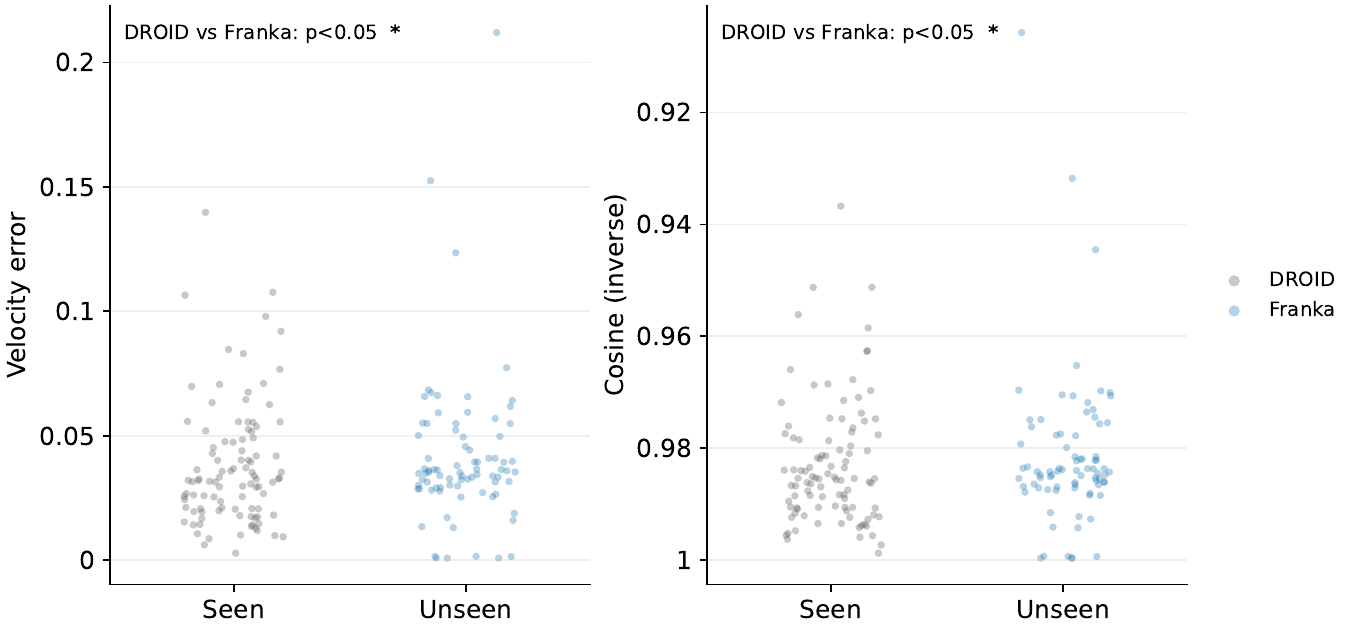}
    \end{minipage}
    \hfill
    \begin{minipage}{0.49\textwidth}
    \subcaption{DreamZero, Relative Joint-Position}
        \centering
        \includegraphics[width=\linewidth]{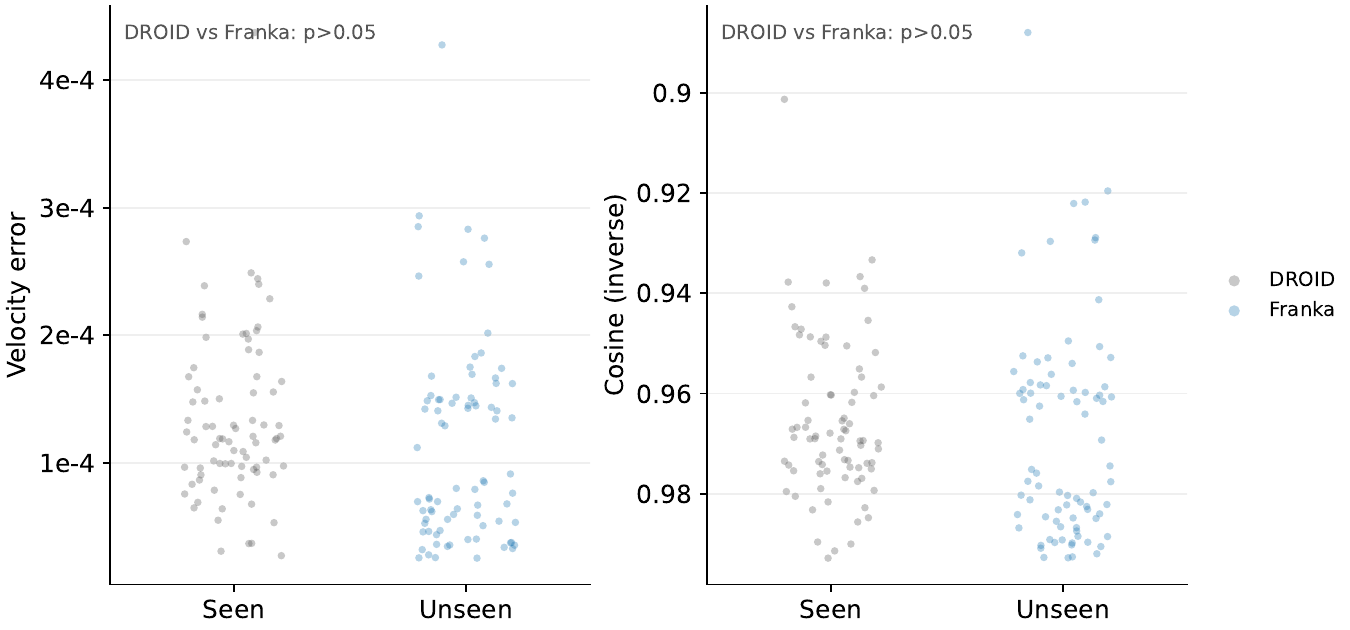}
    \end{minipage}
    \caption{FlowEval results across various models and action representations on the reconstructed franka dataset.}
    \label{appendix:fig:franka}
\end{figure}

\begin{figure}[htbp]
    \centering
    \includegraphics[width=0.8\textwidth]{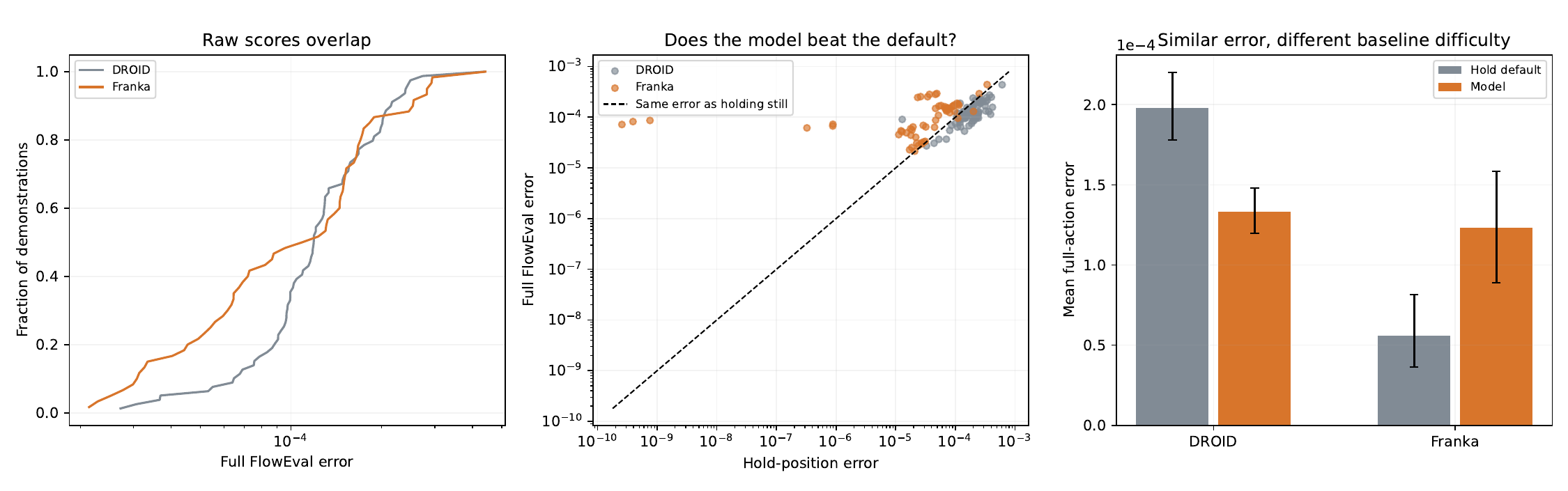}
    \caption{DreamZero comparison with hold still baseline on Franka and DROID.}
    \label{appendix:fig:dreanzero-diagnosis}
\end{figure}

We observe exactly this failure mode. On the reconstructed Franka trajectories, the Franka action targets are concentrated close to the hold-position action, causing the DROID reference and Franka deployment to overlap much more strongly than for the other unseen deployments. DreamZero shows no significant separation ($p>0.05$), while GR00T-N1.6 shows only weaker separation ($p<0.05$), compared with $p<0.001$ on the other unseen target deployments. Upon further investigation (Fig.~\ref{appendix:fig:dreanzero-diagnosis}), we find that DreamZero's mean full-action error is $1.2313\times10^{-4}$, whereas simply predicting the hold-position action gives a lower error of $5.6037\times10^{-5}$. The paired difference is $6.7093\times10^{-5}$, with a task-bootstrap 95\% interval of $[4.1594,9.4574]\times10^{-5}$. 

This failure mode is specific to post-hoc reconstruction of the action labels. In ordinary robot-learning pipelines, converting demonstrations to another control interface would typically require recollecting actions under the desired controller, rather than substituting the next measured robot state as the action target. We therefore do not view this result as limiting the use of \textsc{FlowEval} on ordinary deployment adaptations. Instead, it shows that \textsc{FlowEval} should not be applied to post-reconstructed action targets without verifying that the reconstruction preserves the demonstrated control signal. More broadly, this example also illustrates why post-hoc reconstructed targets of this form can be unsuitable supervision for robot learning: they can collapse toward nearly stationary achieved states even when the original demonstration contains meaningful control commands.

\subsection{Predicting unseen deployments}
\label{appendix:sec:ood}

Seen deployments always have the same embodiment and action representation as one of the pretrained checkpoints. If an unseen deployment has the same embodiment and action representation as one of the pretrained checkpoints already support, we select the corresponding checkpoint and action representation space (Table. ~\ref{appendix:tab:action_channels}). If an unseen deployment does not match any of the configurations, we use the action representation mapping rule specified in Appendix. ~\ref{appendix:part-1-configuration}.

\begin{table}[htbp]
\centering
\caption{Action channels the checkpoints are pretrained on \citep{physical_intelligence2025pi05, fang2026molmoact2, gr00tn1_2025, ye2026dreamzero}.}
\label{appendix:tab:action_channels}
\begin{tabular}{ll}
\toprule
\textbf{Checkpoint} & \textbf{Action Channel} \\
\midrule
DreamZero-DROID        & Relative joint position \\
GR00T-N1.6-DROID       & Relative joint position \\
GR00T-N1.5-DROID       & Absolute joint position; joint velocity \\
GR00T-N1.5             & Delta Cartesian \\
Pi-0.5-Base            & Delta joint position \\
Pi-0.5-DROID           & Joint velocity \\
Pi-0.5-RoboCasa        & Delta Cartesian \\
MolmoAct2              & Mixed \\
MolmoAct2-DROID        & Absolute joint position \\
MolmoAct2-LIBERO       & Delta Cartesian \\
\bottomrule
\end{tabular}
\end{table}

\begin{figure}[htbp]
    \centering
    \begin{minipage}{0.49\textwidth}
    \subcaption{Pi-05-Libero}
        \centering
        \includegraphics[width=\linewidth]{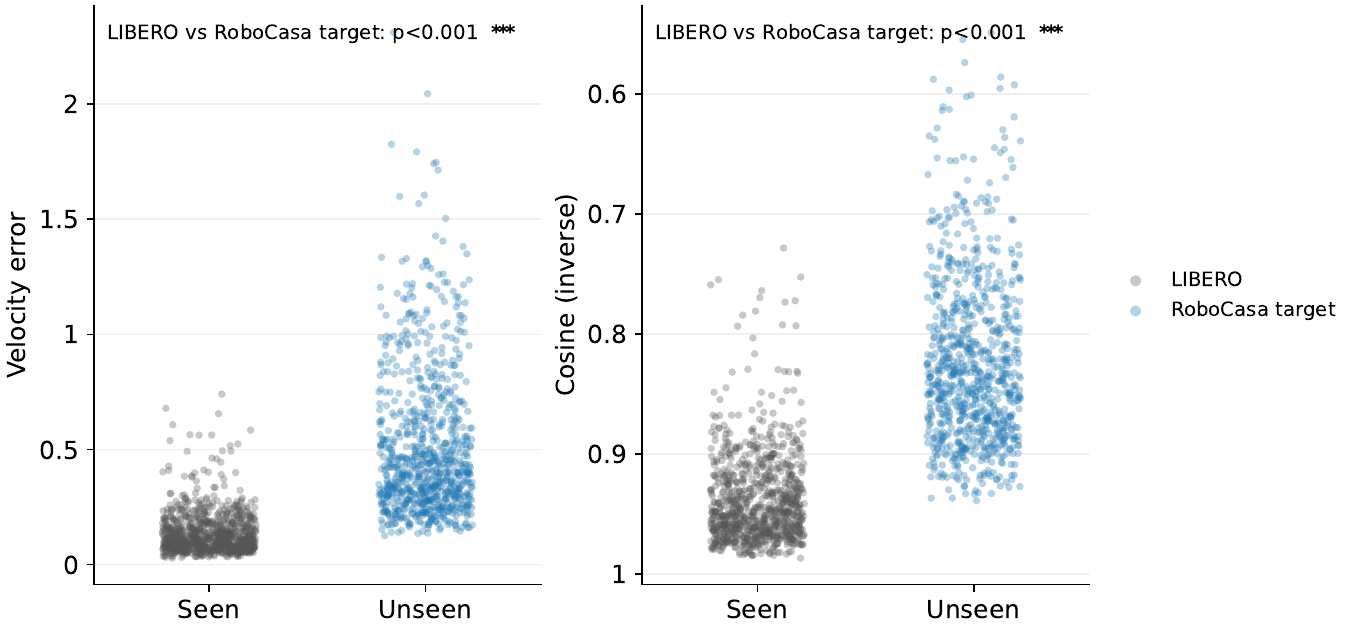}
    \end{minipage}
    \hfill
    \begin{minipage}{0.49\textwidth}
        \subcaption{Pi-05-Robocasa}
        \centering
        \includegraphics[width=\linewidth]{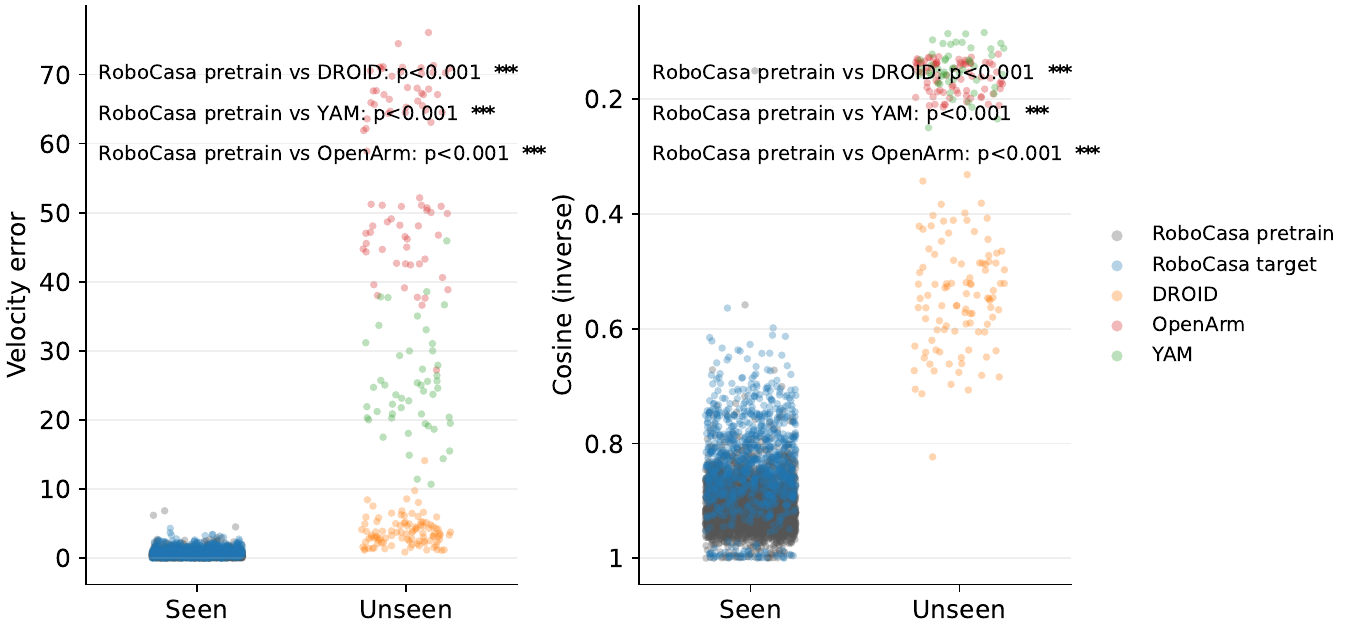}
    \end{minipage}

    \begin{minipage}{0.49\textwidth}
        \subcaption{MolmoAct2-DROID}
        \centering
        \includegraphics[width=\linewidth]{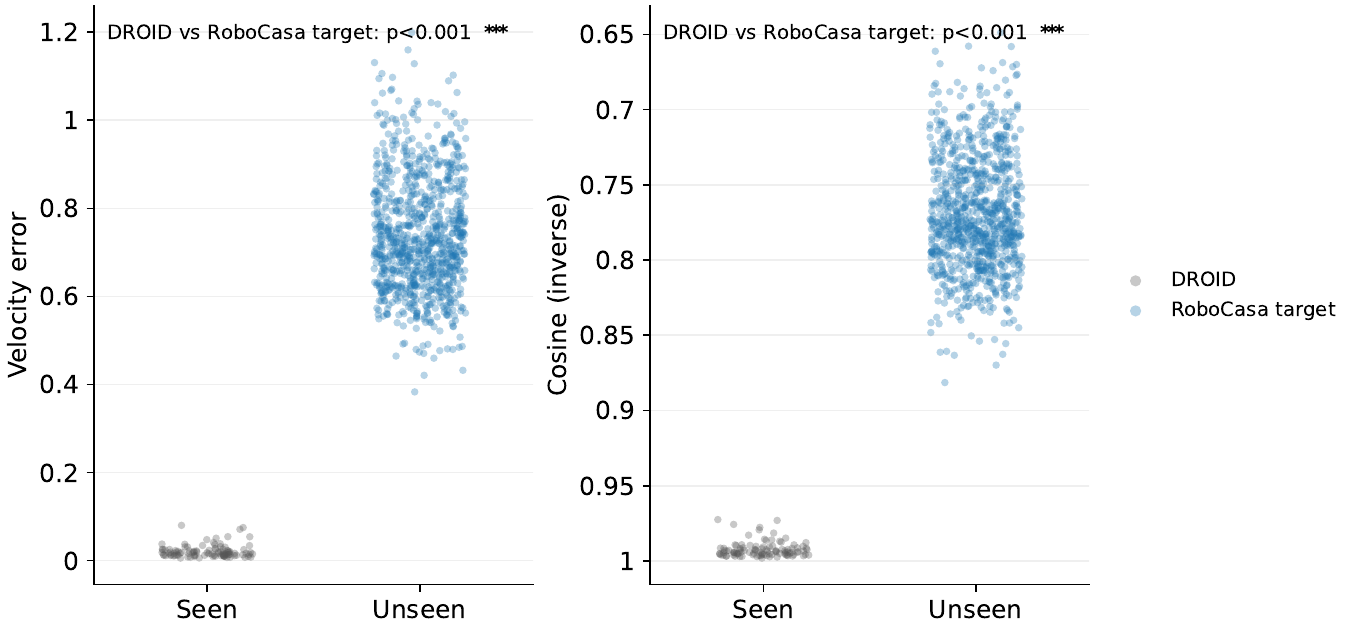}
    \end{minipage}
    \hfill
        \begin{minipage}{0.49\textwidth}
        \subcaption{MolmoAct2-Libero}
        \centering
        \includegraphics[width=\linewidth]{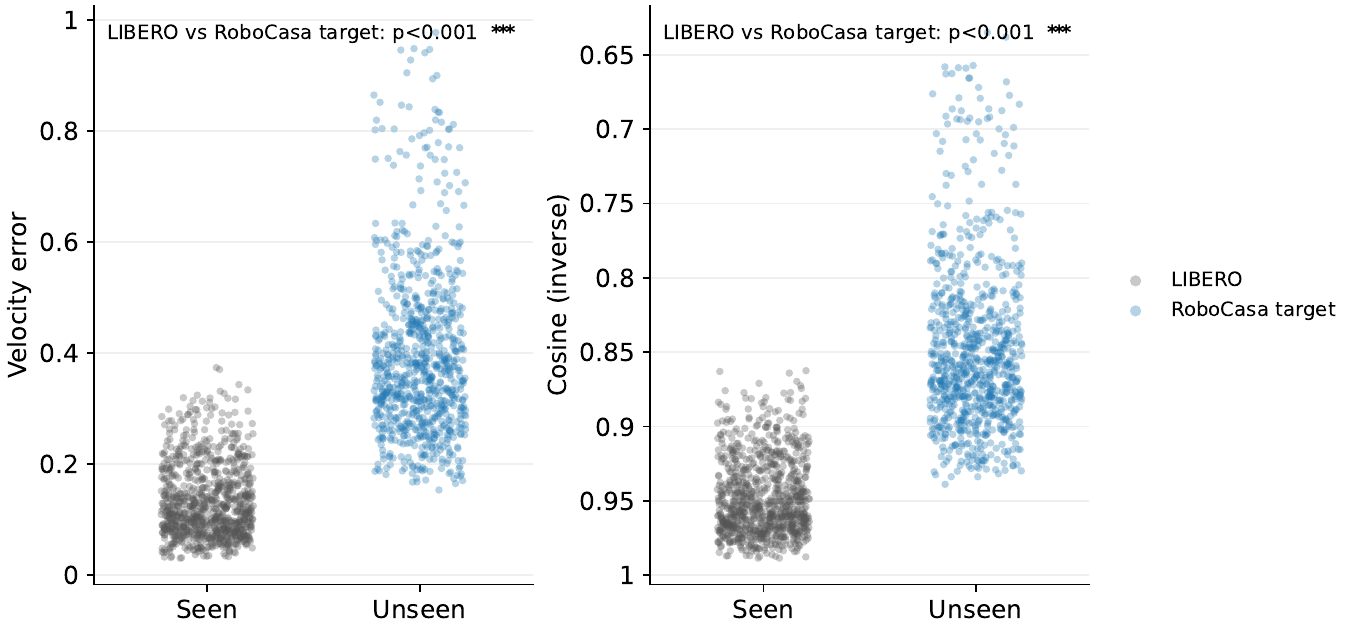}
    \end{minipage}

    \begin{minipage}{0.49\textwidth}
        \subcaption{MolmoAct2}
        \centering
        \includegraphics[width=\linewidth]{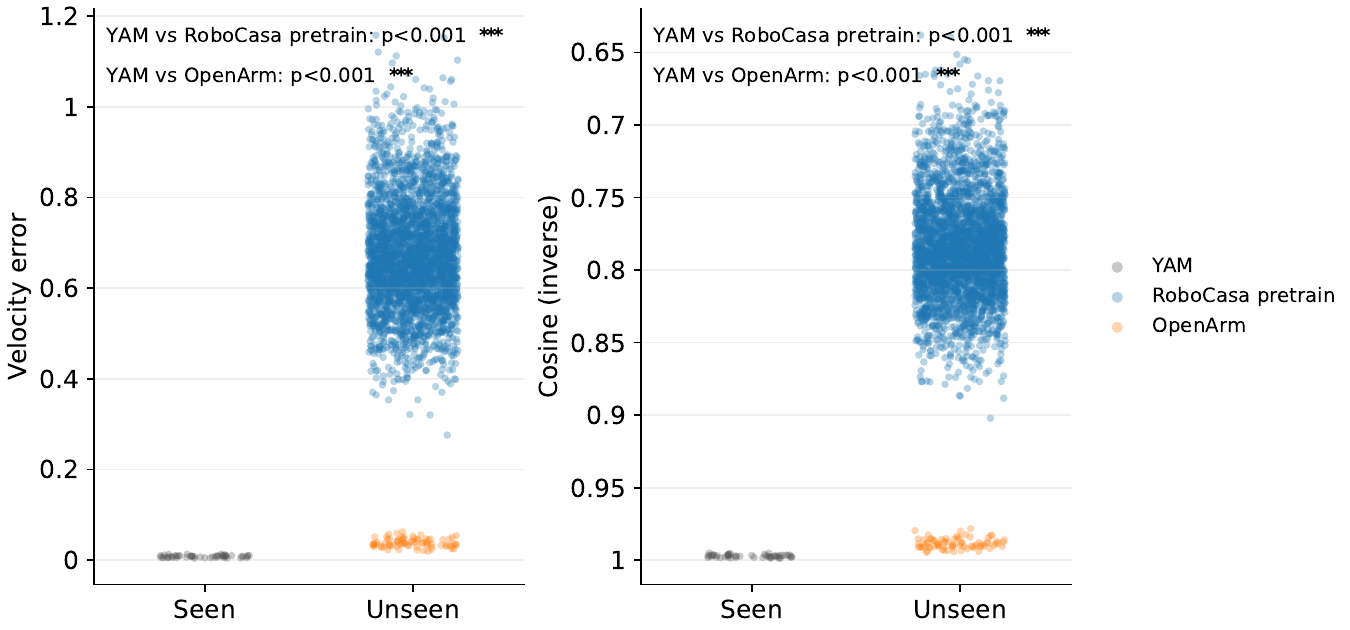}
    \end{minipage}
    \hfill
        \begin{minipage}{0.49\textwidth}
        \subcaption{GR00T-N1.6}
        \centering
        \includegraphics[width=\linewidth]{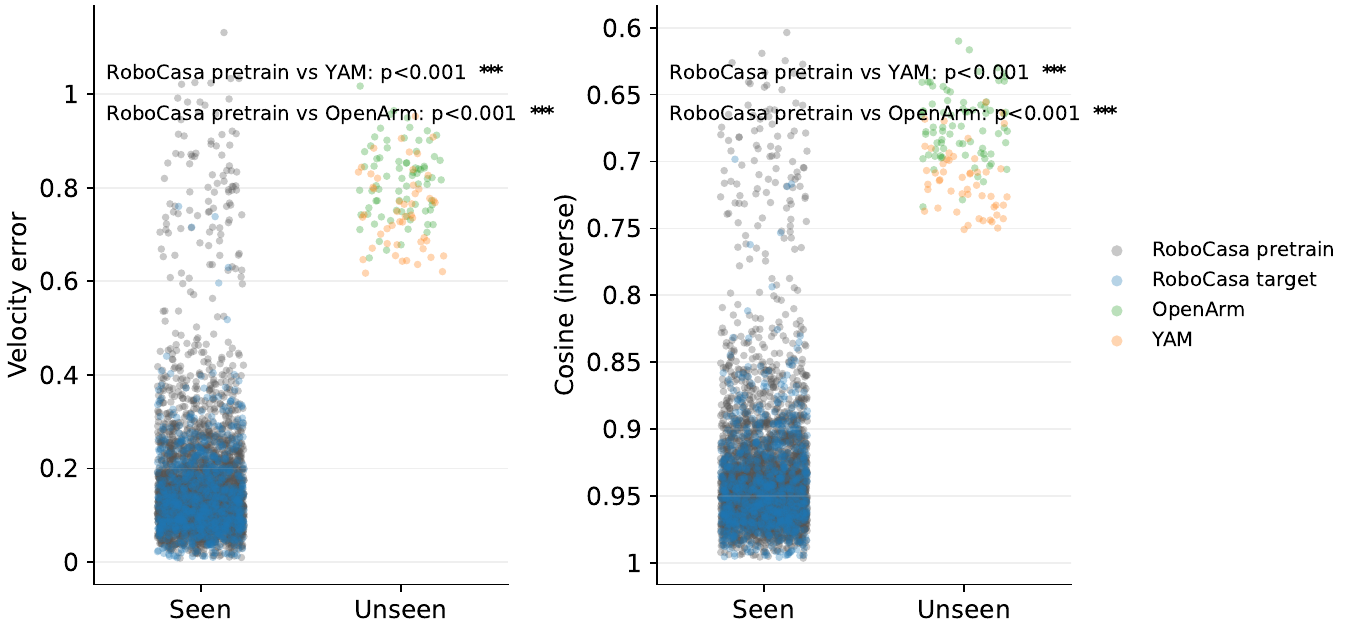}
    \end{minipage}

    \begin{minipage}{0.49\textwidth}
        \subcaption{GR00T-N1.6-DROID}
        \centering
        \includegraphics[width=\linewidth]{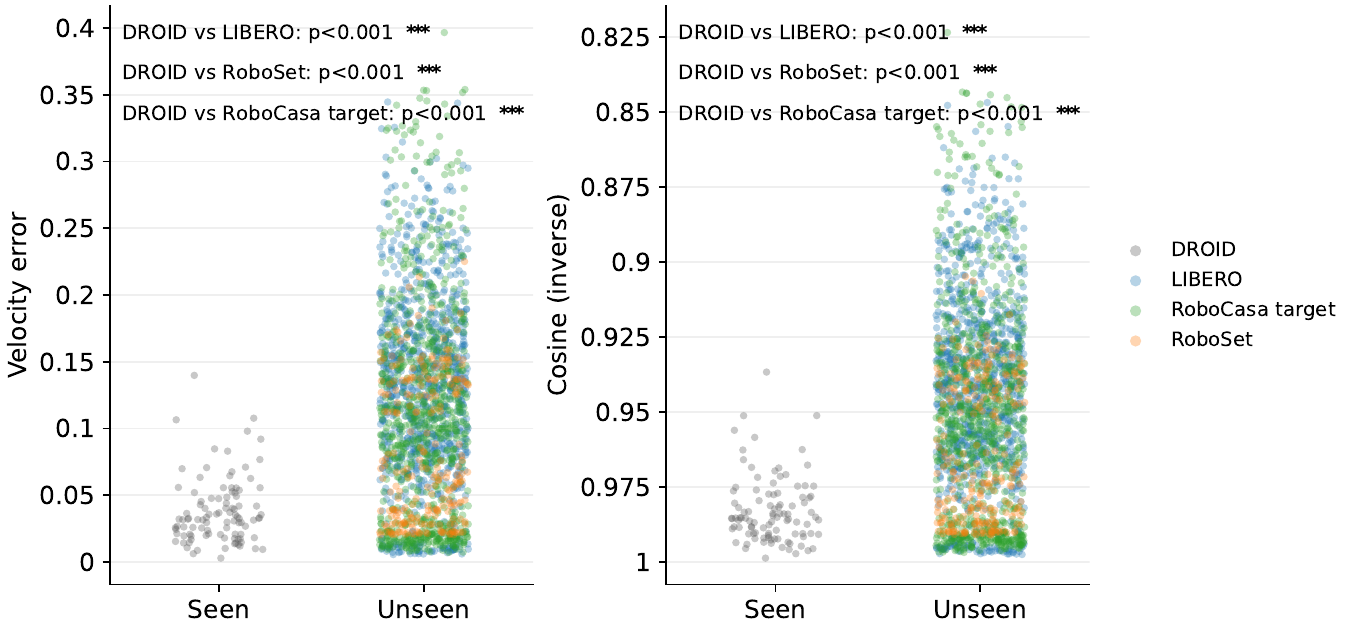}
    \end{minipage}
    \hfill
    \begin{minipage}{0.49\textwidth}
        \subcaption{DreamZero-Droid}
        \centering
        \includegraphics[width=\linewidth]{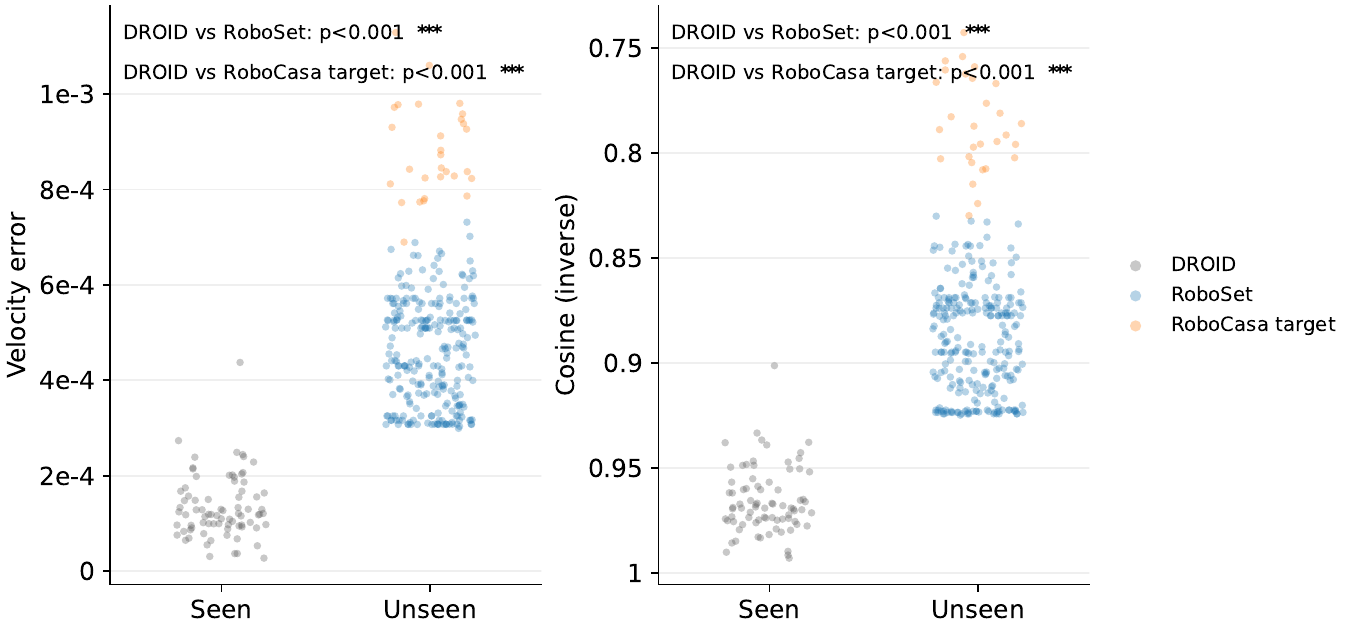}
    \end{minipage}

    \caption{Additional results of \textsc{FlowEval} rating unseen deployment settings.}
    \label{appendix:fig:part1-additional}
\end{figure}

\subsubsection{Statistical tests}
\label{appendix:part-1-stats}

Results for both matched and mismatched action representation spaces are shown in ~\ref{appendix:fig:part1-additional}. The matched condition is a subset reported in the main paper (Sec. ~\ref{sec:ood}). In all cases, unseen deployments are statistically distinguishable ($p<0.001$). We report the p values of the one-sided two-sample Kolmogorov–Smirnov (KS) test in Table.~\ref{appendix:tab:ks_seen_unseen}.

\begin{table*}[t]
\centering
\caption{
Two-sample Kolmogorov--Smirnov tests comparing \textsc{FlowEval} score
distributions on seen-reference and unseen datasets.
All tested unseen deployments are statistically distinguishable from their
corresponding seen-reference distributions.
}
\label{appendix:tab:ks_seen_unseen}
\small
\setlength{\tabcolsep}{6pt}
\begin{tabular}{lllrll}
\toprule
\textbf{Checkpoint} &
\textbf{Seen reference} &
\textbf{Unseen deployment} &
\textbf{KS statistic} &
\textbf{$p$-value} &
\textbf{Sig.} \\
\midrule

\multicolumn{6}{l}{\textbf{Velocity error}} \\
\midrule
Pi-0.5-RoboCasa
    & RoboCasa-pretrain & DROID            & 0.940 & $3.88\times10^{-114}$ & *** \\
Pi-0.5-RoboCasa
    & RoboCasa-pretrain & MolmoAct2-YAM    & 1.000 & $2.57\times10^{-112}$ & *** \\
Pi-0.5-RoboCasa
    & RoboCasa-pretrain & OpenArm          & 1.000 & $4.05\times10^{-162}$ & *** \\
Pi-0.5-LIBERO
    & LIBERO            & RoboCasa-target  & 0.775 & $6.08\times10^{-229}$ & *** \\
MolmoAct2
    & MolmoAct2-YAM     & RoboCasa-pretrain & 1.000 & $5.24\times10^{-112}$ & *** \\
MolmoAct2
    & MolmoAct2-YAM     & OpenArm          & 1.000 & $2.42\times10^{-36}$ & *** \\
MolmoAct2-DROID
    & DROID             & RoboCasa-target  & 1.000 & $1.57\times10^{-140}$ & *** \\
MolmoAct2-LIBERO
    & LIBERO            & RoboCasa-target  & 0.825 & $1.29\times10^{-259}$ & *** \\
GR00T-N1.6
    & RoboCasa-pretrain & MolmoAct2-YAM    & 0.971 & $8.80\times10^{-73}$ & *** \\
GR00T-N1.6
    & RoboCasa-pretrain & OpenArm          & 0.973 & $2.40\times10^{-110}$ & *** \\
GR00T-N1.6-DROID
    & DROID             & LIBERO           & 0.743 & $7.24\times10^{-51}$ & *** \\
GR00T-N1.6-DROID
    & DROID             & RoboSet          & 0.438 & $3.93\times10^{-13}$ & *** \\
GR00T-N1.6-DROID
    & DROID             & RoboCasa-target  & 0.600 & $2.76\times10^{-31}$ & *** \\
DreamZero-DROID
    & DROID             & RoboSet          & 0.987 & $3.00\times10^{-73}$ & *** \\
DreamZero-DROID
    & DROID             & RoboCasa-target  & 1.000 & $5.97\times10^{-27}$ & *** \\

\midrule
\multicolumn{6}{l}{\textbf{Cosine similarity}} \\
\midrule
Pi-0.5-RoboCasa
    & RoboCasa-pretrain & RoboCasa-target  & 0.566 & $9.89\times10^{-198}$ & *** \\
Pi-0.5-RoboCasa
    & RoboCasa-pretrain & DROID            & 0.988 & $1.23\times10^{-161}$ & *** \\
Pi-0.5-RoboCasa
    & RoboCasa-pretrain & MolmoAct2-YAM    & 1.000 & $1.31\times10^{-110}$ & *** \\
Pi-0.5-RoboCasa
    & RoboCasa-pretrain & OpenArm          & 1.000 & $3.24\times10^{-160}$ & *** \\
Pi-0.5-LIBERO
    & LIBERO            & RoboCasa-target  & 0.779 & $2.95\times10^{-231}$ & *** \\
MolmoAct2
    & MolmoAct2-YAM     & RoboCasa-pretrain & 1.000 & $5.24\times10^{-112}$ & *** \\
MolmoAct2
    & MolmoAct2-YAM     & OpenArm          & 1.000 & $2.42\times10^{-36}$ & *** \\
MolmoAct2-DROID
    & DROID             & RoboCasa-target  & 1.000 & $1.57\times10^{-140}$ & *** \\
MolmoAct2-LIBERO
    & LIBERO            & RoboCasa-target  & 0.828 & $1.26\times10^{-261}$ & *** \\
GR00T-N1.6
    & RoboCasa-pretrain & MolmoAct2-YAM    & 0.974 & $2.60\times10^{-74}$ & *** \\
GR00T-N1.6
    & RoboCasa-pretrain & OpenArm          & 0.976 & $3.27\times10^{-113}$ & *** \\
GR00T-N1.6-DROID
    & DROID             & LIBERO           & 0.737 & $8.57\times10^{-50}$ & *** \\
GR00T-N1.6-DROID
    & DROID             & RoboSet          & 0.416 & $7.02\times10^{-12}$ & *** \\
GR00T-N1.6-DROID
    & DROID             & RoboCasa-target  & 0.593 & $1.68\times10^{-30}$ & *** \\
DreamZero-DROID
    & DROID             & RoboSet          & 0.987 & $3.00\times10^{-73}$ & *** \\
DreamZero-DROID
    & DROID             & RoboCasa-target  & 1.000 & $5.97\times10^{-27}$ & *** \\

\bottomrule
\end{tabular}
\end{table*}

\subsection{Action Representation Mapping}
\label{appendix:part-1-configuration}

When comparing a pretrained model with a deployment setting that uses a different action representation, the model and dataset dimensions may not align directly. In such cases, we follow the mapping rules below.

Some mappings may appear unintuitive because the paired dimensions do not share the same physical units. However, the goal of \textsc{FlowEval} is to probe the adaptation gap between the model prediction and the downstream ground truth. For example, if a model is pretrained to predict joint positions while the downstream objective uses joint velocities, adaptation must still learn the corresponding distribution shift. We therefore pair joint positions with joint velocities in such cases.

Semantic mapping is used to identify the lowest-friction correspondence for learning this distribution shift. It does not eliminate or override the shift itself; rather, it provides the simplest correspondence through which that shift can be learned. Accordingly, gripper dimensions are mapped to gripper dimensions, arm dimensions to arm dimensions, and base dimensions to base dimensions.

1) Valid dataset dimensions determine the dimensions involved in the loss computation. 

\begin{itemize}
    \item This dimension in the dataset should be valid and non-trivial. Examples include joint positions or gripper states. Counter-examples include dimensions that remain a constant, such as the torso lift dimension in robocasa.
    \item Each valid dataset dimension require a corresponding model output dimension. When len(dataset) $>$ len(valid\_model\_output), where valid\_model\_output means the dimensions in the model that's activated (actually trained with much data), always take the 'unused' next model output dimensions to fill in the slot. When len(dataset) $<$ len(valid\_model\_output), discard some model outputs so that they map one by one.
\end{itemize}

2) Same semantic dimensions should map to each other.
\begin{itemize}
    \item Grippers should be mapped with grippers, arms should be mapped with arms, and bases should be mapped with bases.
    \item For bimanual to single arm mapping, map the single arm to one of the bimanual arms.
\end{itemize}

3) Arms should be mapped to arms.
\begin{itemize}
\item Dataset dimensions decide the dimensions involved: If dataset has 6D arm and the model has 7D arm, discard the last arm dim in model and only include 6 dims in the loss computation; if dataset jas 7D arm and the model has 6D arm, use the next 'unused' dimension (note that this may not be the immediate following dimension, since the model may have other outputs e.g. base/gripper after the arm. It should be the next 'unused' not associated with a semantic under the config) to map with the last arm dim in the dataset and include 7 dims in the loss computation.
\end{itemize}

\subsection{Predicting downstream adaptation}
\subsubsection{Control 1: Deployment shift}
\label{appendix:sec:deployment-shift-control}

In Sec.~\ref{sec:regression} we observe significant interactions between success rate gain and how much additional data is provided, indicating that \textsc{FlowEval} provides useful signal about how performance in an unseen deployment scales with additional data. This control examines whether this signal reflects a generic scaling effect or is specific to deployment shift.
To this end, we additionally finetune Pi05-RoboCasa and GR00T-N1.6 (without shift) and compare their results with Pi05-base and GR00T-N1.5 (with shift) on the RoboCasa tasks. All models use 50, 150, and 500 demonstrations except GR00T-N1.5, for which we use 50 and 150 (Appendix.~\ref{appendix:fig:collapse}). 

\begin{table}[ht]
\centering
\small
\begin{tabular}{lllrr}
\hline
Shift & Model & Demos & Cosine similarity & Velocity error / loss \\
\hline
With & Pi05-base & 50/150/500
& $\mathbf{+0.474}$ ($\mathbf{.0047}$)
& $\mathbf{-0.0382}$ ($\mathbf{.0093}$) \\
With & GR00T-N1.5 & 50/150
& $\mathbf{-5.537}$ ($\mathbf{1.81\times10^{-8}}$)
& $\mathbf{+2.363}$ ($\mathbf{3.12\times10^{-8}}$) \\
Without & Pi05-RoboCasa & 50/150/500
& $+0.275$ ($.3112$)
& $-0.0382$ ($.1056$) \\
Without & GR00T-N1.6 & 50/150/500
& $+0.266$ ($.4521$)
& $-0.1102$ ($.4464$) \\
\hline
\end{tabular}
\caption{Initial-metric $\times$ demonstration-budget interaction
coefficients, with $p$-values in parentheses. Bold denotes $p<.05$.}
\label{tab:deployment-shift-control}
\end{table}

Both metrics show significant interactions with demonstration budget
under deployment shift, whereas neither is significant without shift.
These results support the interpretation that the interaction arises
from deployment shift rather than a generic scaling effect.

\begin{table}[htbp]
\centering
\small
\begin{tabular}{llrr}
\hline
Deployment shift & Model & Cosine similarity & Velocity error / loss \\
\hline
With    & Pi05-base     & $\mathbf{+0.474}$ ($\mathbf{.0047}$)
                       & $\mathbf{-0.0382}$ ($\mathbf{.0093}$) \\
With    & GR00T-N1.5    & $-0.0073$ ($.9964$)
                       & $-0.0258$ ($.9699$) \\
Without & Pi05-RoboCasa & $+0.275$ ($.3112$)
                       & $-0.0382$ ($.1056$) \\
Without & GR00T-N1.6    & $+0.266$ ($.4521$)
                       & $-0.1102$ ($.4464$) \\
\hline
\end{tabular}
\caption{Initial-metric $\times$ demonstration-budget interaction
coefficients, with $p$-values in parentheses; bold denotes $p<.05$.
Positive cosine coefficients associate higher initial cosine similarity
with larger gains; negative loss coefficients associate lower initial
loss with larger gains.}
\label{tab:deployment-shift-control}
\end{table}

\subsubsection{Control 2: Shared Task Effects}
\label{appendix:sec:task-control}

One alternative explanation for the observed interaction is shared task effects instead of
model--deployment fit. If some tasks are consistently easier to learn, and these differences in task learnability align with \textsc{FlowEval} scores, they could induce a correlation with scaling. In short, we want to distinguish whether \textsc{FlowEval} measures a merely task level effect or a model-deployment interaction effect.

We test this possibility through cross-model task consistency. 
We compare Pi05-Base, Pi05-RoboCasa, GR00T-N1.5, and GR00T-N1.6. We compute
pairwise Pearson correlations across shared tasks, with Holm correction across
the six model pairs for each outcome. We additionally test task effects on
success after accounting for model using a partial F-test, and test cross-model
metric consistency using Bartlett's test of sphericity.

\textsc{FlowEval} scores show limited evidence of cross-model task consistency. No pairwise cosine-similarity correlations remain significant after Holm correction, and only one velocity-error correlation does, between Pi05-Base and Pi05-RoboCasa ($r=0.634$, adjusted $p=0.0374$). Bartlett's tests do not reject identity for either metric's cross-model correlation matrix ($p=0.0854$ for cosine similarity and $p=0.0945$ for velocity error). These results provide little evidence that \textsc{FlowEval} consistently ranks the same tasks similarly across models, a pattern compatible with model--task-specific variation. Therefore, \textsc{FlowEval} reasonably measures the model-deployment interaction effect we hypothesized.

\begin{table*}[t]
\centering
\small
\setlength{\tabcolsep}{6pt}
\caption{Cross-model task consistency of success and FlowEval metrics. Strong
success consistency confirms shared task effects in RoboCasa; corresponding
FlowEval consistency would indicate that these effects could induce the
FlowEval--performance relationship.}
\label{appendix:tab:task-consistency-control}
\begin{tabular}{lll}
\toprule
 & Cross-model pairwise results & Global test \\
\midrule
Cosine similarity
& 0/6 pairs after Holm correction
& Bartlett $p=0.0854$ \\
Velocity error
& 1/6 pairs: Pi05-Base/Pi05-RoboCasa, $r=0.634$, adjusted $p=0.0374$
& Bartlett $p=0.0945$ \\
\bottomrule
\end{tabular}
\end{table*}

\subsubsection{Control 3: Headroom bias}
\label{appendix:sec:headroom-control}

Another alternative explanation for the observed interaction is baseline success. Because
success is bounded at 100\%, a task starting at 20\% success has more room to
improve than a task starting at 90\%. If \textsc{FlowEval} is correlated with starting performance, part of its interaction with demonstration count could arise from this difference in available headroom rather than from an association
with scaling beyond starting performance. 

We therefore control for each task's measured success at 50 demonstrations and,
critically, allow the scaling slope itself to depend on this baseline success.
This asks whether \textsc{FlowEval} remains associated with scaling after accounting for the possibility that tasks with different starting success rates have different rates of improvement.

\paragraph{Control specification.}
For each task, let $R$ denote success rate at a given demonstration count,
$L=\log(N)$ the natural logarithm of the number of demonstrations, $m$ the
task-level step-zero FlowEval metric, and $S_{50}$ the task's success rate with
50 demonstrations. The original regression is
\begin{equation}
    R = \beta_0 + \beta_1 L + \beta_2 m + \beta_3 Lm + \epsilon,
    \label{eq:original-scaling}
\end{equation}
where $\beta_3$ is the FlowEval metric $\times$ log-demonstration-count
interaction examined in Section~4.2.

The baseline-success-controlled regression is
\begin{equation}
    R = \beta_0 + \beta_1 L + \beta_2 m + \beta_3 Lm
        + \beta_4 S_{50} + \beta_5 L S_{50} + \epsilon.
    \label{eq:headroom-control}
\end{equation}
The $S_{50}$ term accounts for differences associated with starting success,
while $L S_{50}$ allows the relationship between demonstration count and
success to vary with starting success. Including only $S_{50}$ would therefore
not directly address the alternative explanation for the scaling interaction.
The comparison of interest is how the FlowEval interaction $\beta_3$ changes
after adding both terms.

$L$, $m$, and $S_{50}$ are centered at their cell-level means. The original and
controlled regressions use identical observations. No headroom-normalized
gain or separate gain regression is introduced, so this analysis directly
tests the original interaction under an additional baseline-success control.

\begin{table*}[t]
\centering
\small
\setlength{\tabcolsep}{5pt}
\caption{Control for baseline success in the FlowEval scaling analysis. Reported coefficients are the FlowEval metric $\times \log(\mathrm{demos})$ interaction $\beta_3$. The controlled specification additionally includes 50-demo success $S_{50}$ and its interaction with $\log(\mathrm{demos})$. Coefficient magnitudes should be compared before and after adjustment within each metric rather than across metrics.}
\label{appendix:tab:headroom-control}
\begin{tabular}{lllccc}
\toprule
Model & Demos & Metric & $\beta_3$ & 95\% CI & $p$ \\
\midrule
Pi05-Base & 50/150/500 & Cosine similarity (Original)   & $0.4738$   & $[0.1455,\,0.8022]$       & $0.00468$ \\
           &            & Cosine similarity (Controlled) & $0.2595$   & $[-0.04683,\,0.5659]$     & $0.09684$ \\
Pi05-Base & 50/150/500 & Velocity error (Original)     & $-0.03820$ & $[-0.06699,\,-0.00942]$   & $0.009283$ \\
           &            & Velocity error (Controlled)   & $-0.02106$ & $[-0.04391,\,0.001801]$   & $0.07099$ \\
\midrule
GR00T-N1.5 & 50/150 & Cosine similarity (Original)   & $-5.537$ & $[-7.465,\,-3.609]$ & $1.81{\times}10^{-8}$ \\
            &        & Cosine similarity (Controlled) & $-5.468$ & $[-7.563,\,-3.373]$ & $3.12{\times}10^{-7}$ \\
GR00T-N1.5 & 50/150 & Velocity error (Original)     & $2.363$ & $[1.526,\,3.200]$ & $3.12{\times}10^{-8}$ \\
            &        & Velocity error (Controlled)   & $2.337$ & $[1.433,\,3.241]$ & $4.07{\times}10^{-7}$ \\
\bottomrule
\end{tabular}
\end{table*}

\paragraph{Results.}
For Pi05-Base, controlling for baseline success reduces the cosine and velocity-error interactions by 45.2\% and 44.9\%, respectively, while preserving their directions and approximately 55\% of their original magnitudes. The negative $L S_{50}$ coefficients ($p=0.00184$--$0.00435$) indicate that tasks with higher baseline success tend to gain less from additional demonstrations. This suggests that a ceiling effect contributes to the observed scaling trend but does not fully explain it.

For GR00T-N1.5 over 50--150 demonstrations, controlling for baseline
success leaves the cosine and velocity-error interactions almost unchanged.
The cosine coefficient changes from $-5.537$ to $-5.468$ (1.24\%), and the
velocity-error coefficient changes from $2.363$ to $2.337$ (1.11\%). 

\subsubsection{Training Configurations}
\label{appendix:sec:training-configs}

We use these training configurations for Pi05-Base, Pi05-Robocasa, GR00T-N1.5, and GR00T-N1.6.

\textbf{OpenPi Hyperparameters}

\begin{verbatim}
global_batch_size: 64
train_steps: 30000

optimizer: AdamW
beta1: 0.9
beta2: 0.95
epsilon: 1e-8
weight_decay: 1e-6
grad_clip_global_norm: 1.0

lr_schedule: warmup-cosine
warmup_steps: 1000
peak_lr: 2.5e-5
decay_steps: 30000
end_lr: 2.5e-6

ema_decay: 0.99
\end{verbatim}

\textbf{GR00T Hyperparameters}

\begin{verbatim}
learning_rate: 1e-4
weight_decay: 1e-5
warmup_ratio: 0.05

global_batch_size: 64
gradient_accumulation_steps: 1

shard_size: 1024
num_shards_per_epoch: 100000
episode_sampling_rate: 0.1

optim: adamw_torch

tune_llm: true
tune_visual: true
tune_projector: true
tune_diffusion_model: true

state_dropout_prob: 0.0
random_rotation_angle: null
extra_augmentation_config: null

load_bf16: false
reproject_vision: false
eagle_collator: true
backbone_trainable_params_fp32: true
use_relative_action: true
\end{verbatim}

\subsubsection{Training Collapse in GR00T-N1.5 with 500 Demonstrations}
\label{appendix:sec:collapse}

\begin{figure}[htbp]
    \centering
    \includegraphics[width=\linewidth]{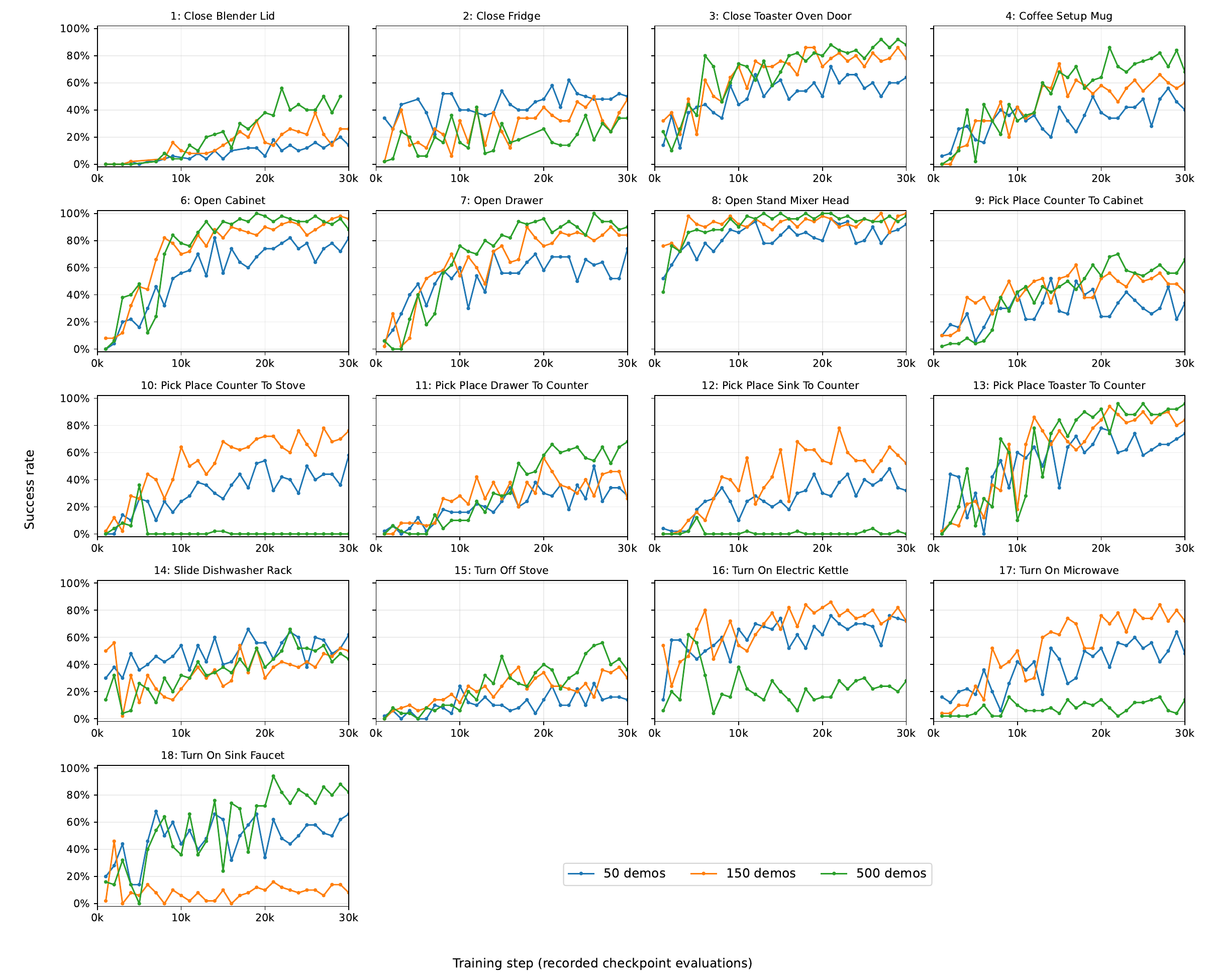}
    \caption{Performance curves for GR00T-N1.5 finetuning. Training collapse occurs on task 18 with 150 demonstrations and on tasks 2, 10, 12, 16, and 17 with 500 demonstrations.}
    \label{appendix:fig:collapse}
\end{figure}

\textbf{Loss Spikes.} We train each GR00T-N1.5 finetuning run for
30k steps to allow sufficient optimization. Around 5k steps, when the
learning-rate schedule plateaus, we observe loss spikes and occasional
training instability. Performance collapse affects 5 of 17 runs
(29.4\%) at 500 demonstrations (Fig.~\ref{appendix:fig:collapse}),
compared with one GR00T-N1.5 run at 150 demonstrations and one or two
runs across all demonstration counts for the Pi-05 variants
(Table~\ref{appendix:tab:collapse}). We therefore exclude the
500-demonstration condition from the primary GR00T-N1.5 regression
analysis and report the interaction over 50--150 demonstrations.

\begin{table}[ht]
\centering
\small
\caption{Runs exhibiting persistent or sudden performance collapse.}
\label{appendix:tab:collapse}
\begin{tabular}{llcc}
\hline
Model & Demonstrations & Collapsed runs & Fraction of runs \\
\hline
GR00T-N1.5     & 500 & 5 & \textbf{29.4\%} (5/17) \\
GR00T-N1.5     & 150 & 1 & 5.88\% (1/17) \\
GR00T-N1.6 (Action Expert)     & All & 0 & 0.00\% (0/51) \\
Pi-05-Base     & All & 1 & 1.96\% (1/51) \\
Pi-05-RoboCasa & All & 2 & 3.92\% (2/51) \\
\hline
\end{tabular}
\end{table}

\end{document}